\documentclass{article}

\usepackage{microtype}
\usepackage{graphicx}
\usepackage{subfigure}
\usepackage{booktabs} 

\usepackage{hyperref}

\usepackage[accepted]{icml2025}

\usepackage{amsmath}
\usepackage{amssymb}
\usepackage{mathtools}
\usepackage{amsthm}

\usepackage{graphicx}
\usepackage{booktabs}
\usepackage{multirow}
\usepackage{color}  
\usepackage{colortbl}
\usepackage{booktabs}
\usepackage{amssymb}
\usepackage{makecell}
\usepackage{subcaption}
\usepackage{algorithm}
\usepackage{algorithmic}
\usepackage{newfloat}
\usepackage{listings}
\usepackage{alltt}
\usepackage{xcolor}
\definecolor{ForestGreen}{RGB}{34,139,34}

\usepackage[capitalize,noabbrev]{cleveref}

\theoremstyle{plain}

\theoremstyle{definition}

\theoremstyle{remark}

\usepackage[textsize=tiny]{todonotes}

\begin{document}

\icmltitlerunning{Asymmetric Decision-Making in Online Knowledge Distillation}

\twocolumn[
\icmltitle{Asymmetric Decision-Making in Online Knowledge Distillation: \\Unifying Consensus and Divergence}



\icmlsetsymbol{equal}{*}

\begin{icmlauthorlist}
\icmlauthor{Zhaowei Chen}{equal,jiiov}
\icmlauthor{Borui Zhao}{equal,jiiov}
\icmlauthor{Yuchen Ge}{sch}
\icmlauthor{Yuhao Chen}{jiiov}
\icmlauthor{Renjie Song}{jiiov}
\icmlauthor{Jiajun Liang}{jiiov}
\end{icmlauthorlist}

\icmlaffiliation{jiiov}{JIIOV Technology}
\icmlaffiliation{sch}{University of Southern California}

\icmlcorrespondingauthor{Zhaowei Chen}{chaowechan@gmail.com}
\icmlcorrespondingauthor{Jiajun Liang}{tracyliang18@gmail.com}

\icmlkeywords{Machine Learning, ICML}

\vskip 0.3in
]



\printAffiliationsAndNotice{\icmlEqualContribution} 

\begin{abstract}

Online Knowledge Distillation (OKD) methods streamline the distillation training process into a single stage, eliminating the need for knowledge transfer from a pretrained teacher network to a more compact student network. This paper presents an innovative approach to leverage intermediate spatial representations. Our analysis of the intermediate features from both teacher and student models reveals two pivotal insights: (1) the similar features between students and teachers are predominantly focused on foreground objects. (2) teacher models emphasize foreground objects more than students. Building on these findings, we propose Asymmetric Decision-Making (ADM) to enhance feature consensus learning for student models while continuously promoting feature diversity in teacher models. Specifically, Consensus Learning for student models prioritizes spatial features with high consensus relative to teacher models. Conversely, Divergence Learning for teacher models highlights spatial features with lower similarity compared to student models, indicating superior performance by teacher models in these regions. Consequently, ADM facilitates the student models to catch up with the feature learning process of the teacher models. Extensive experiments demonstrate that ADM consistently surpasses existing OKD methods across various online knowledge distillation settings and also achieves superior results when applied to offline knowledge distillation, semantic segmentation and diffusion distillation tasks.
\end{abstract}

\section{Introduction}


In conventional offline knowledge distillation paradigms \cite{hinton2015distilling, tian2019contrastive, chen2021distilling}, a pre-trained teacher model provides soft supervision to guide student training through post-hoc knowledge transfer. This two-stage paradigm inherently suffers from computational redundancy and temporal decoupling between teacher-student interactions. In contrast, recently proposed Online Knowledge Distillation (OKD) methods simplify to one-stage distillation where both teacher and student models learn simultaneously from scratch, while achieving competitive or even superior performance \cite{guo2020online, qian2022switchable, song2023multi, zhang2023generalization, zhang2018deep}. Typically, Deep Mutual Learning (DML) \cite{zhang2018deep} encourages each network to mutually learn from each other by matching their output logit via KL divergence. In this paper, we propose a novel Asymmetric Decision-Making strategy that unifies consensus and divergence feature learning to achieve better OKD performance.

\begin{figure*}[!ht]
    \centering
    \includegraphics[scale=0.75]{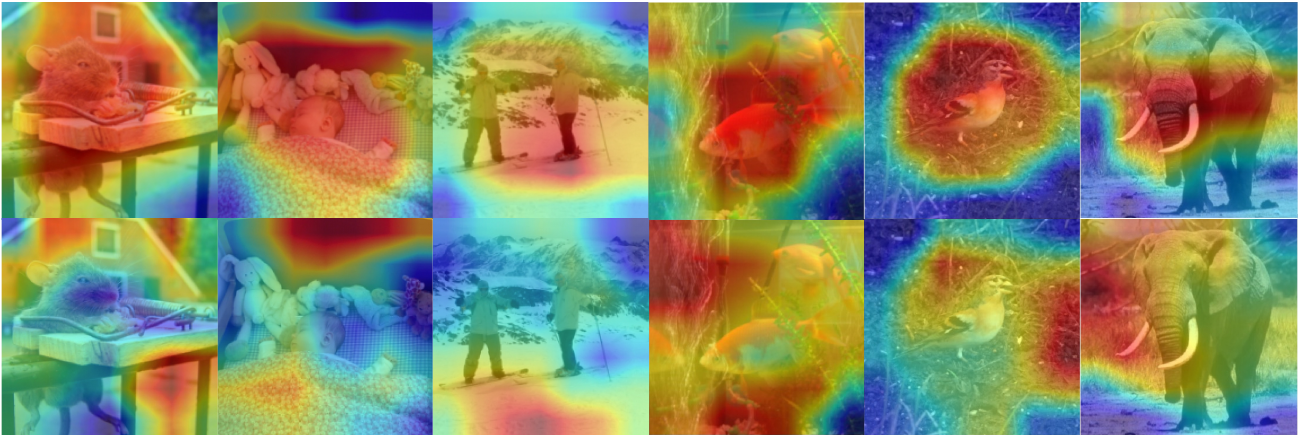}  
    \caption{\textbf{Observation of Discrepancy Regions between Teacher and Student Features.} The first row shows that discrepancy regions between teacher and student's features are more concentrated on foreground object regions after Vanilla training. In contrast, the second row displays discrepancy regions are mainly found in background regions after ADM training. Best viewed in color.}
    \vspace{-10pt}
    \label{fig:2}
\end{figure*}

\begin{figure}[!h]
    \centering
    \subfigure[mIoU of CAM regions.]{\includegraphics[scale=0.28]{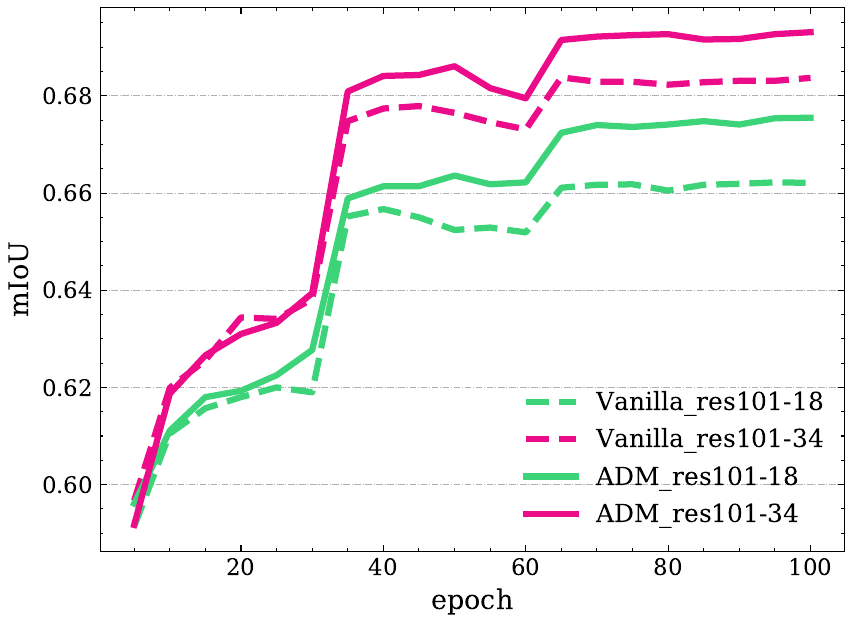}}
    \subfigure[mIoU of similar and discrepancy regions.]{\includegraphics[scale=0.28]{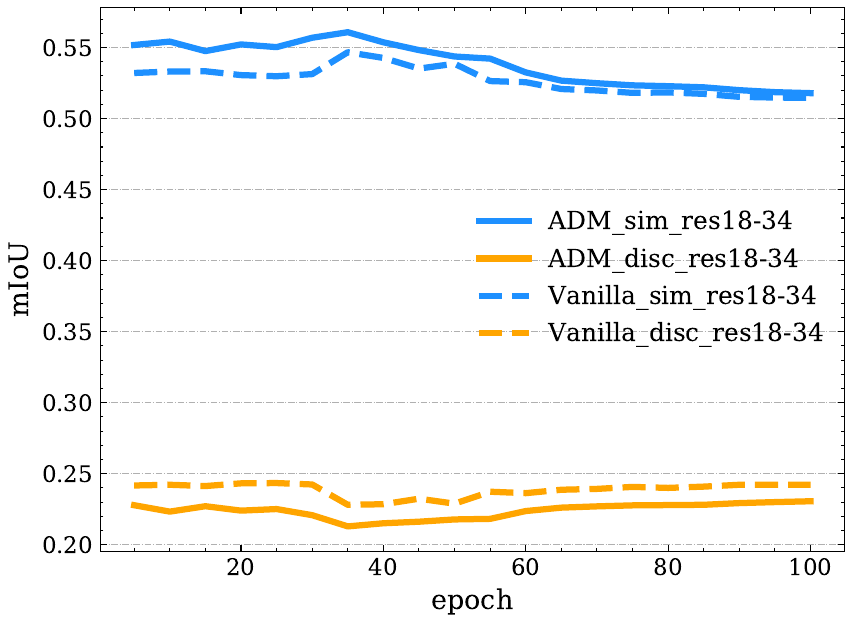}}

    \caption{\textbf{Analysis of the Teacher and Student Models' Intermediate Features.} We load a pre-trained ResNet101 model and obtain its CAM as the Target Object Regions (dubbed TOR) on ImageNet. (a) \textbf{mIoU of CAM regions.} We calculate the mean intersection over union (mIoU) between TOR and the CAM of intermediate ResNet-34, ResNet-18 models saved during the training process. We observe that ResNet-34 is more concentrated on object regions than ResNet-18. (b) \textbf{mIoU of similar and discrepancy feature regions.} We observe that the similar regions between ResNet-34 and ResNet-18 predominantly focus on TOR and a substantial proportion of discrepancy regions are also located within the TOR. }
    \label{fig:CAM_vis}
\vspace{-15pt}
\end{figure}


This paper investigates a fundamental question: How does intermediate feature alignment affect model behavior in online distillation? Through systematic analysis using Class Activation Mapping(CAM) \cite{zhou2016learning} (Figure~\ref{fig:2} and~\ref{fig:CAM_vis}), we reveal two pivotal findings: (1) \textbf{the similar features between student and teacher models are predominantly concentrated on foreground objects}, suggesting students prioritize learning these "easy" patterns. (2) \textbf{teacher models highlight foreground objects more than students}, indicating an underutilized capacity for discriminative feature discovery. Surprisingly, the most informative divergent features (which may explain teacher superiority) still concentrate on foreground areas rather than background contexts.



Building on these insights, we propose Asymmetric Decision-Making (ADM). It is a unified framework that coordinates consensus enhancement and divergence exploration through spatially-aware feature modulation. Unlike conventional symmetric distillation, ADM implements role-specific learning strategies: For students, we amplify feature attribution in teacher-aligned foreground regions to accelerate confidence building. For teachers, we intensify exploration of under-activated foreground patterns to avoid premature convergence to student-level representations. This strategic asymmetry creates a dynamic where students progressively close the performance gap while teachers continuously unveil new discriminative features. Our main contributions can be summarised as follows.
\begin{itemize}
\item Through interpretability-driven analysis, we identify the paradoxical coexistence of foreground-centric feature consensus and teacher-student attribution gaps, challenging conventional assumptions about knowledge transfer in online settings.


\item We propose a simple but effective learning strategy named ADM, which allocates different attention to feature maps based on spatial similarity. ADM encourages student models to boost the attribution scores on foreground objects while reinforce the strengths of teacher models continually.

\item Comprehensive experiments are conducted to verify our effectiveness, including online KD, offline KD, semantic segmentation and diffusion distillation. It is shown that the proposed method consistently outperforms state-of-the-art OKD methods.
\end{itemize}



\section{Related Work}

\textbf{Knowledge Distillation} KD aims to transfer knowledge acquired by an effective but cumbersome teacher model to a smaller compact student model. KD has attracted wide interest in vision and language applications \cite{meng2022distillation, niu2022respecting, sanh2019distilbert}. KD can be formulated in logits-based \cite{hinton2015distilling, zhao2022decoupled, huang2022knowledge, sun2024logit}, feature-based \cite{heo2019comprehensive, chen2021distilling, li2022self, xiaolong2023norm, huang2024knowledge}, and relation-based \cite{tian2019contrastive}. Recent studies have focused on automating the process of discovering effective knowledge distillation strategies or architectures \cite{li2024kd, li2023automated, dong2023diswot, hao2024one}. Besides, for the foreground and background imbalance in object detection, FGD \cite{yang2022focal} proposed to use ground-truth boxes to separate the images and guide the student to focus on crucial pixels and channels through spatial and channel attention. GID \cite{dai2021general} select the discriminative instances with higher scores for distillation in object detection. These two methods under comparison primarily rely on bounding boxes to explicitly discern the foreground and background for distillation, which markedly contrasts with our proposed method. 

\textbf{Online Knowledge Distillation} The traditional offline KD framework needs a pre-trained and fixed teacher model. Online KD methods are more attractive because the training process is simplified to a single stage \cite{li2022distilling, li2021online, lin2022tree, yang2023online}. Deep Mutual Learning (DML) \cite{zhang2018deep} enables students to share knowledge from each other’s predictions to achieve distillation without teachers. KDCL \cite{guo2020online} integrates the output of multiple students under different data augmentations as soft labels to guide students to optimize. SwitOKD \cite{qian2022switchable} attempts to close the accuracy gap by slowing down the training of the teacher during training. SHAKE \cite{li2022shadow} bridges offline and online knowledge transfer by building an extra shadow head as a proxy teacher model to perform mutual distillation with the student model. Recently proposed online KD methods mainly focus on logits-based. In this work, ADM unifies consensus and divergence feature learning for teacher and student models, which leverages the spatial similarity between teacher and student models, thus implicitly distinguishing between the foreground and background. Moreover, the foreground and background in our method embody a crucial attribute, representing simple and hard regions respectively. Consequently, our approach facilitates the student model to learn from the teacher model progressively, following a gradient of difficulty that ranges from easy to challenging tasks. This constitutes an innovative perspective that has not previously been proposed in the existing literature.

\textbf{Diffusion Distillation} Diffusion distillation methodologies can be taxonomically categorized into step distillation and model distillation. \cite{salimans2022progressive, meng2023distillation} established the iterative progressive distillation framework, compressing DDPM sampling steps from 1,000+ to 4 through cyclic teacher-student optimization. The DreamFusion paradigm \cite{poole2022dreamfusion} first bridged text-to-3D generation through differential rendering and SDS losses. Recent work \cite{nguyen2024swiftbrush, yin2024one} achieves 40× acceleration via adaptive step annealing. Adversatial Distillation \cite{sauer2025adversarial} integrated adversarial loss to ensure high image fidelity even in the low-step regime of one or two sampling steps. \cite{kim2025bk} proposed block pruning and feature distillation for low-cost general-purpose Text-to-image (T2I) generation. In this paper, we integrate ADM into both step distillation and model distillation to verify its effectiveness.


\begin{figure*}[t]
    \centering
    \includegraphics[scale=0.6]{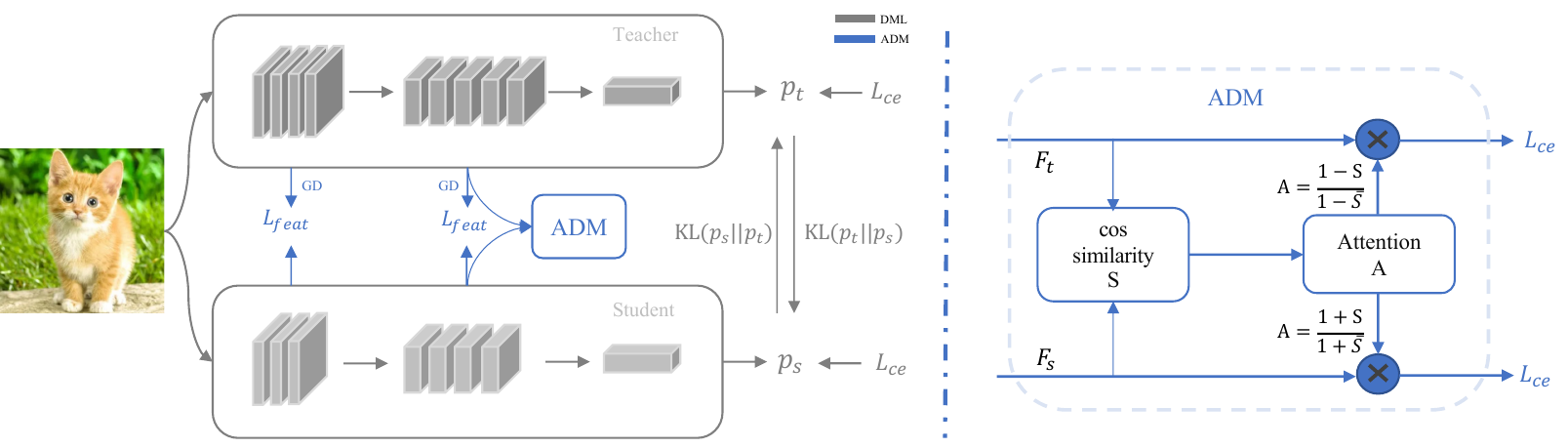}  
    \caption{\textbf{Asymmetric Decision-Making in Online Knowledge Distillation.} GD means Gradient Detach operation. Consensus Learning for student models and Divergence Learning for teacher models. The gray arrow represents the baseline of the DML method, while the blue arrow denotes the newly added ADM component. Best viewed in color.}
    \label{fig:3}
    \vspace{-10pt}
\end{figure*}

\vspace{-5pt}

\section{Preliminary}
\label{sec:Preliminary}

\subsection{Online Knowledge Distillation}

Generally, OKD replaces the commonly used pre-trained teachers with peer-trainable models. The training loss consists of the Cross-Entropy (CE) loss and the Knowledge Distillation (KD) loss.
Let $\mathcal{D}=\{x_i, y_i\}_{i=1}^{N}$ be a training set containing $N$ images and $C$ categories of labels. $y_i \in \{1, \cdots, C\}$ is the ground-truth label of the image $x_i$. The $m$-th model ($m \in \{1, \cdots, M\}$) obtains its output logits $\mathbf{z}^m = f(x;\theta ^m) \in \mathcal{R^C}$, where $\theta ^m$ is the training parameters of the model $m$.
The classification loss is calculated by Cross-Entropy:
\setlength{\abovedisplayskip}{2pt} 
\setlength{\belowdisplayskip}{2pt} 
\begin{equation}
\label{eq:ce}
    {\mathcal{L}_{ce}}\left({\mathbf{z}^m}, y_i\right) = - \log \frac {\exp \left({\mathbf{z}^m_{y_i}}\right)} {\sum_{c=1}^{C} \exp \left({\mathbf{z}_c^{m}}\right)},
\end{equation}

The KD loss usually minimizes the KL divergence between the probabilistic outputs of different models:
\begin{equation}
\label{eq:kd}
    {\mathcal{L}}_{kd} = \tau^2 \mathcal{D}_{KL} \left(p^m,p^n\right)=\tau^2 \sum_{c=1}^{C}{p^n_c \log\frac{p^n_c}{p^m_c}},
\end{equation}
where $p^m, p^n \in \mathcal{R^C}$ are the soft logits produced by models $m$ and $n$.
The soft logits are calculated by:
\vspace{5pt}
\begin{equation}
\label{eq:soft_logits}
    p^m=\sigma({{\mathbf{z}}^m}/{\tau})=\frac{\exp({{\mathbf{z}}^m}/{\tau})}{\sum_c {\exp({{\mathbf{z}}^m_c}/{\tau})} },
\end{equation}
\vspace{10pt}
where $\sigma$ is the softmax function, and $\tau$ is the temperature factor to control the softness of logits. Then the DML loss is formulated as follows and $\lambda$ is the trade-off weight:
\vspace{5pt}
\begin{equation}
\label{eq:dml}
    {\mathcal{L}_{dml}} = \sum_{i=1}^{M} ({\mathcal{L}_{ce}^i} + \lambda {\mathcal{L}_{kd}^i}),
\end{equation}

\vspace{5pt}
\subsection{Online Feature Distillation}
\label{eq:feat}
In recent offline KD methods, FitNet\cite{romero2014fitnets} combines the feature-based loss with the logit-based loss to improve the performance. In online KD, we observed that making the teacher mimic the student's features would compromise the diversity of the teacher's features, thereby leading to detrimental effects on performance. Due to the accelerated convergence and higher accuracy of teacher models, teachers exhibit superior features that effectively facilitate the learning process for students. Thereby, we simply use the following MSE loss for the student feature distillation but detach the gradient of teacher feature distillation:
\vspace{3pt}
\begin{equation}
    \mathcal{L}_{feat} = \frac{1}{B}\sum_{b=1}^B\frac{1}{HWC} \sum_{i=1}^H\sum_{j=1}^W\sum_{c=1}^C(F_{bijc}^t - \phi(F_{bijc}^s))^2,
\end{equation}

where $B$ is the batch size. $i$, $j$ are the location of corresponding feature map with width $W$ and height $H$. $C$ represents the number of channels.
$F^t$ and $F^s$ are the intermediate features of student and teacher models, respectively. The function $\phi$ is the 1$\times$1 convolutional adaptation function to adapt $F^s$ to the same dimension as $F^t$.

\section{Method}
\label{sec:Method}
\subsection{Asymmetric Decision-Making in Online Knowledge Distillation}
\label{Consensus and Divergence Learning}

In this section, we show how to design Asymmetric Decision-Making in Online Knowledge Distillation. Intuitively, during the training, the teacher's object-centric features should be strengthened while the student features should be changed accordingly. The above analysis suggests that ADM function can be decomposed into two components, namely, Consensus loss $\mathcal{L}_{co}$ and Divergence loss $\mathcal{L}_{di}$. $\mathcal{L}_{co}$ is applied to student model, indicating that the higher similarity between the student and teacher models in specific regions, the stronger the supervision we provide for those regions. The similarity matrix is described as:

\begin{equation}
     \label{eq:cos_sim}
\mathcal{S}_{ij}=\mathcal{D}(F^t_{ij}, F^s_{ij})=\frac{F^t_{ij} {F^s_{ij}}^T}{\left\|F^t_{ij}\right\|_2\left\|F^s_{ij}\right\|_2},
\end{equation}

where $F^t, F^s \in \mathcal{R}^{h \times w \times d}$, $\mathcal{S} \in \mathcal{R}^{h \times w}$, $d$ is the channel number. $i$ and $j$ are the location of the corresponding feature map with width $w$ and height $h$. As shown in Figure \ref{fig:3}. For simplicity, we only take the penultimate feature before Global Average Pooling (GAP) as input. $\mathcal{L}_{co}$ encourages the student to focus on learning simple knowledge rather than concentrating on difficult regions where a large discrepancy with the teacher model, described as:

\begin{equation}
     \label{eq: loss_co}
     \mathcal{L}_{co} = {\mathcal{L}_{ce}}\left(W^s\left(\psi_{avg}\left (\frac{1+\mathcal{S}}{1+\Bar{\mathcal{S}}} \cdot F^s\right)\right), y_i\right),
\end{equation}

where $W^s$ is the parameters of the last student linear classifier. $\psi_{avg}$ is GAP operation. $\Bar{\mathcal{S}}$ is the mean value of $\mathcal{S}$ along the dimension of $h$ and $w$. The Attention Function for students is a monotone-increasing function. Simultaneously, to avoid the teacher model learning an excessively high similarity with the student model and falling into a local optimum, we introduce $\mathcal{L}_{di}$ to the teacher model, allowing it to continuously strengthen the learning of discrepancy regions from the student model, described as:

\begin{equation}
     \label{eq: loss_ad}
     \mathcal{L}_{di} = {\mathcal{L}_{ce}}\left(W^t\left(\psi_{avg}\left (\frac{1-\mathcal{S}}{1-\Bar{\mathcal{S}}} \cdot F^t\right)\right), y_i\right),
 \end{equation}
 
where $W^t$ is the parameters of the last teacher linear classifier. The Attention Function for teachers is a monotone-decreasing function. By summing up both losses as follows, the teacher model constantly guides and drives the student model to learn and catch up. $\alpha$ and $\beta$ are hyper-parameters. We here employ CE loss function in ADM for conciseness, and in addition, as shown in Section \ref{sec: different cal loss}, we have investigated the combination of CE loss with KD loss, which achieved competitive results.


\begin{equation}
     \label{eq: decompose CAL}
     \mathcal{L}_{adm} = \alpha \mathcal{L}_{co} + \beta \mathcal{L}_{di},
 \end{equation}

As a result, the overall training loss $\mathcal{L}_{total}$ can be composed of DML loss, feature loss, and adm loss:
\begin{equation}
     \label{eq:total loss}
     \mathcal{L}_{total} = {\mathcal{L}_{dml}} + \gamma {\mathcal{L}_{feat}} + {\mathcal{L}_{adm}} ,
 \end{equation}
 
where $\gamma$ is a factor for balancing the losses. In this way, via the ADM loss, we have enabled the student to match the teacher network's feature adaptively, thus boosting OKD performance to a great extent.


\section{Experiment} 
\label{sec:Implementation}
\subsection{Experimental settings}
\textbf{Datasets.} We validated ADM on the following datasets: 
\textbf{CIFAR-10} \cite{krizhevsky2009learning} consists of 60000 32x32 colour images in 10 classes, with 6000 images per class. 
\textbf{CIFAR-100} \cite{krizhevsky2009learning} consists of 32$\times$32 color images drawn from 100 classes, which are split into 50k train and 10k test images. \textbf{ImageNet} \cite{russakovsky2015imagenet} is a large scale image classificaiton dataset that contains 1k classes with about 1.28 million training images and 50k images for validation. Each image is resized to 224x224. On ImageNet and CIFAR-100, we adopt the commonly used random crop and horizontal flip techniques for data augmentation.
\textbf{Cityscapes} \cite{cordts2016cityscapes} is a dataset for urban scene parsing, comprising 5000 meticulously annotated images. The distribution of these images is as follows: 2975 for training, 500 for validation, and 1525 for testing.

\textbf{Teacher-Student Pairs.}
To validate the generalizability of different distillation methods, we select a group of popular network architectures to form different teacher-student pairs. By combining different teacher and student networks, like ResNets \cite{he2016deep}, WideResNet(WRN)s \cite{zagoruyko2016wide}, VGGs \cite{simonyan2014very}, MobileNets \cite{howard2017mobilenets, sandler2018mobilenetv2}, and ShuffleNets \cite{ma2018shufflenet, zhang2018shufflenet}, we can perform distillation between the same architectures (e.g., ResNet34 \cite{he2016deep}-ResNet18\cite{he2016deep}) and different architectures (e.g., ResNet50\cite{he2016deep}-MobileNet-V2\cite{sandler2018mobilenetv2}). ADM introduces no additional trainable parameters. The added computational cost is markedly minimal relative to the original. For example, in the ResNet34-18 experiments on ImageNet, ResNet34 has 3.67G FLOPs and ResNet18 has 1.82G FLOPs. ADM only adds 0.52M FLOPs.

\textbf{Implementation Details.} 
Following the settings of previous methods \cite{qian2022switchable, tian2019contrastive}, the batch size, epochs, learning rate decay rate, and weight decay rate are 256/128, 100/300, 0.1/0.1, and 0.0001/0.0005, respectively on ImageNet/CIFAR-100. The initial learning rate is 0.1 on ImageNet, and 0.01 for MobileNetV2, 0.1 for the other students on CIFAR-100. Besides, the learning rate drops every 30 epochs on ImageNet and drops at 140, 200, 250 epochs on CIFAR-100. The optimizer is Stochastic Gradient Descent (SGD) with momentum 0.9. By following the conventions in OKD, we use a fixed temperature T as 1.0 and loss weight $\lambda$ as 1.0 for all experiments. For a fair comparison, the hyper-parameters of different methods are fixed in all experiments. We set $\alpha$ = 0.01, $\beta$ = 0.01, $\gamma$ = 1.0 on CIFAR-100 dataset because of limited spatial information, and $\alpha$ = 0.2, $\beta$ = 0.6, $\gamma$ = 0.01 on ImageNet dataset for all OKD experiments. It's worth noting that ADM exhibits no sensitivity to hyperparameter settings as shown in Table \ref{tab:alpha_beta_weight} and \ref{tab:gamma_weight}.

\subsection{Online KD Experiments on Image Classification}

\textbf{Results on CIFAR-100.} Table \ref{tab:cifar100-same} and Table \ref{tab:cifar100-unsame} report the experimental results on CIFAR-100 with similar architecture style and dissimilar architecture style teacher-student pairs, respectively. \textbf{We run all methods five times with different seeds and obtain the average accuracy. }
ADM is a plug-and-play module that can effectively enhance the performance of existing methods. Note that ADM can perform much better, especially for heterogeneous student-teacher pairs, +0.35\% vs +1.41\%. This is because heterogeneous architectures focus on different regions on intermediate features. It is more discriminative for students to leverage the feature similarity compared to output logits. 

\begin{table}[h]
\renewcommand\arraystretch{1.3}
\setlength\tabcolsep{0.4mm}  
\centering
\small

\footnotesize
\resizebox{0.47\textwidth}{!}{%
\begin{tabular}{l|cccccc}
\multicolumn{1}{c|}{Backbone} & Vanilla & DML   & KDCL  & SwitOKD & DML+ADM       & SwitOKD+ADM            \\ \Xhline{3\arrayrulewidth}
ResNet8$\times$4                     & 72.33   & 73.95 & 74.61 & 74.49   & 74.23 & \textbf{74.63} \\
ResNet32$\times$4                    & 79.55   & 79.99 & 79.84 & 79.31   & 79.62        & \textbf{80.01}                 \\ \hline
VGG-8                         & 70.58   & 72.45 & 71.78 & 72.71   & 72.8  & \textbf{72.94} \\
VGG-13                        & 75.12   & 76.19 & 74.63 & 75.64   & 76.07        & \textbf{76.50}                 \\
\end{tabular}
} %

\caption{\textbf{Top-1 Accuracy (\%) comparison on CIFAR-100 with same architecture style teacher-student pairs.} The upper and lower models denote student and teacher, respectively.}
\label{tab:cifar100-same}
\vspace{-10pt}
\end{table}

\begin{table}[h]  
\renewcommand\arraystretch{1.3}
\setlength\tabcolsep{0.6mm}
\centering
\small

\footnotesize
\resizebox{0.44\textwidth}{!}{%
\begin{tabular}{l|cccccc}
\multicolumn{1}{c|}{Backbone} & Vanilla & DML   & KDCL  & SwitOKD & DML+ADM                 \\ \Xhline{3\arrayrulewidth}
0.5MobileNetV2                & 60.39   & 66.69 & 65.98 & 66.68   & \textbf{68.10}   \\
WRN-16-2                      & 73.01   & 73.59 & 71.54 & 73.55   & \textbf{73.62}                           \\ \hline
ShuffleNetV1                  & 68.44   & 73.58 & 73.91 & 73.82   & \textbf{74.61}           \\
ResNet32x4                    & 79.30   & \textbf{80.08} & 78.97 & 79.37   & 79.97                            \\ \hline
ShuffleNetV2                  & 70.17   & 75.15 & 75.00 & 75.18   & \textbf{75.75}   \\
ResNet32$\times$4                    & 79.38   & 80.15 & 79.14 & 79.90   & \textbf{80.26}                            \\
\end{tabular}
}
\caption{\textbf{Top-1 Accuracy (\%) comparison on CIFAR-100 with different architecture style teacher-student pairs.} The upper and lower models denote student and teacher, respectively.}
\label{tab:cifar100-unsame}
\vspace{-5pt}
\end{table}

\textbf{Results on ImageNet.} Table \ref{tab:imagenet} reports the performance of our approach on ImageNet with similar ResNets architectures. We run all methods three times with different seeds and obtain the Top-1 average accuracy. As presented in the table, ADM outperforms other OKD methods in most cases. Equipped with ADM, the DML baseline of ResNet-18 (11.7M parameters) gains 0.47$\%$ in Top-1 accuracy. Besides, we observe that a larger teacher model ResNet-101 (44.5M parameters) fails to distill a better compact student in other OKD methods. But in the paradigm of ADM, due to the teacher's superior performance, more confidence can be set in the discrepancies regions between the teacher and student, which, in turn, students learn uncomplicated features rather than challenging ones. As a result, ADM can effectively bridge the gap between the large teacher and student. In addition, even for two identical models, ADM can still enhance both performances. As shown in Table \ref{tab:imagenet-different}, With pair ResNet18 and 0.5MobileNetV2 (1.97M parameters), ADM obtains 0.89\% accuracy gains than DML, and ResNet50 (25.6M parameters)-MobileNetV1 (4.23M parameters) pair 0.43\% gains, verifying its effectiveness on heterogeneous architectures. It can be seen that \emph{the relative difference in parameter size is not the main reason why a larger model with better accuracy cannot serve as a good teacher for smaller models}. How to choose a good teacher model for distillation remains an open question worth discussing.

It's worth mentioning that all existing OKD methods encountered an identical issue. The performance gain on the homogenous backbone pair is not obvious compared to the heterogenous pair. There are primarily two reasons for this observation. Firstly, heterogeneous pairs exhibit greater structural differences, enabling them to acquire a broader range of knowledge from one another. Secondly, and more significantly, the student model within heterogeneous pairs is smaller. The performance disparity between the student and teacher models is more pronounced, resulting in a larger potential for improvement. An additional experimental observation is that the degradation of the teacher model's performance is a common issue among all existing OKD methods. However, our method outperforms other baseline approaches in terms of the teacher's performance improvement. 



\begin{table}[h]
\vspace{-10pt}
\renewcommand\arraystretch{1.3}
\setlength\tabcolsep{0.6mm}
\centering
\small
\footnotesize
\resizebox{0.42\textwidth}{!}{%
\begin{tabular}{l|ccccc}
\multicolumn{1}{c|}{Backbone}                            & Vanilla                                              & DML                                                  & KDCL           & SwitOKD & DML+ADM                \\ \Xhline{3\arrayrulewidth}
ResNet-18                                                & 70.16                                                & 71.12                                                & 71.54          & 71.00   & \textbf{71.59} \\
ResNet-34                                                & 73.93                                                & 73.78                                                & 74.36          & 73.15   & 73.93                 \\ \hline
ResNet-18                                                & 70.16                                                & 71.44                                       & 71.33          & 71.36   & \textbf{71.62}                 \\
ResNet-101                                               & 76.62                                                & 76.80                                                & 78.22          & 76.28   & 77.81                 \\ \hline
ResNet-50 &  75.99 & 76.70 & 76.70          & 76.78   & \textbf{77.10} \\
ResNet-101                                               & 77.85                                                & 78.17                                                & 78.36          & 78.00   & 78.31                 \\ \hline
ResNet-18 & 70.16                                                & 70.80                                                & \textbf{71.09} & 70.70   & \textbf{71.07} \\
ResNet-18                                                & 70.16                                                & 70.80                                                & 71.08          & 70.63   & 70.94                
\end{tabular}
}
\caption{\textbf{Top-1 Accuracy (\%) comparison on ImageNet with same architecture style teacher-student pairs.} The upper and lower models denote student and teacher, respectively.}
\label{tab:imagenet}
\end{table}

\begin{table}[h]
\vspace{-20pt}
\renewcommand\arraystretch{1.3}
\setlength\tabcolsep{0.6mm}
\centering
\small
\footnotesize
\resizebox{0.45\textwidth}{!}{%
\begin{tabular}{l|ccccc}
\multicolumn{1}{c|}{Backbone} & Vanilla & DML   & KDCL  & SwitOKD & DML+ADM                \\ \Xhline{3\arrayrulewidth}
0.5MobileNetV2                & 62.49   & 64.21 & 64.32 & 63.94   & \textbf{65.10} \\
ResNet-18                     & 70.43   & 69.00 & 70.75 & 68.81   & 69.66                 \\ \hline
MobileNet-V1                  & 68.81   & 71.13 & 70.76 & 71.07   & \textbf{71.69}  \\
ResNet-50                     & 76.38   & 75.44 & 76.42 & 74.80   & 75.68                
\end{tabular}
}
\caption{\textbf{Top-1 Accuracy (\%) comparison on ImageNet with different architecture style teacher-student pairs.} The upper and lower models denote student and teacher, respectively. }
\label{tab:imagenet-different}
\vspace{-10pt}
\end{table}

\subsection{Transfer Learning to Semantic Segmentation}

Experiments are also conducted on semantic segmentation, which is recognized as a demanding dense prediction task. We train DeepLabV3~\cite{chen2018encoder} and PSPNet~\cite{zhao2017pyramid} with ResNet-18 backbone on Cityscapes dataset. Different pre-trained models are considered using an online teacher with ResNet-101 backbone. As the results summarized in Table~\ref{tab:seg}, our ADM can significantly outperform existing OKD methods on semantic segmentation task.

\begin{table}[H]
\renewcommand\arraystretch{1.3}
\setlength\tabcolsep{0.6mm}
\centering
\small
\footnotesize
\resizebox{0.47\textwidth}{!}{%
\begin{tabular}{c|cc|cccc}
Methods   & Teacher & Student & DML & KDCL & SwitOKD & DML+ADM    \\ 
\Xhline{3\arrayrulewidth}
DeepLabV3  & 78.30 & 73.58  & 74.19 & 73.82 & 73.98 & \textbf{74.55}  \\ 
PSPNet     & 76.96 & 70.74  & 71.54 & 71.49 & 71.72 & \textbf{71.98}  \\ 
\end{tabular}
}
\caption{\textbf{Results on Cityscapes val dataset.} All models are pretrained on ImageNet.}
\label{tab:seg}
\vspace{-10pt}
\end{table}

\vspace{-15pt}
\subsection{Extension to Diffusion Distillation}

\begin{figure}[H]
    \centering
    \vspace{-10pt}
    \includegraphics[scale=0.45]{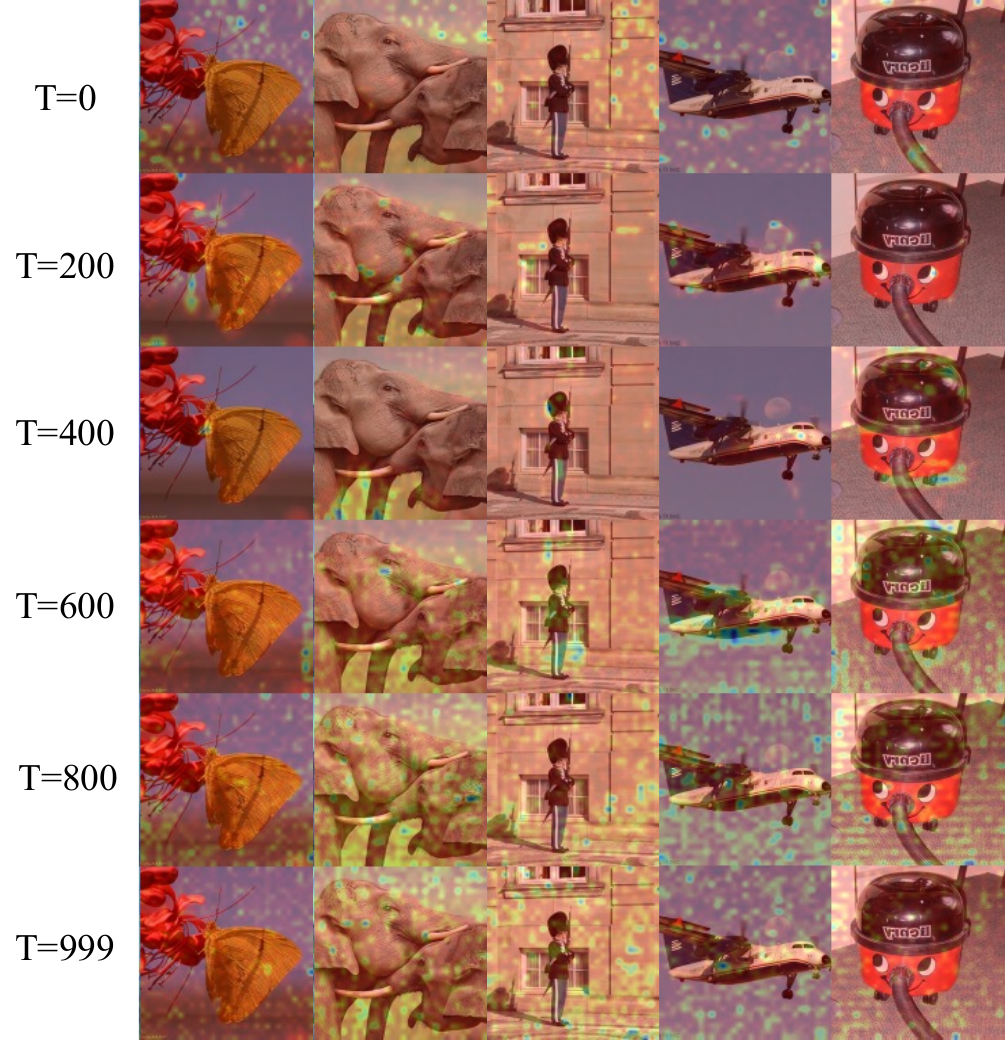}  
    \caption{\textbf{Visualization of the spatial-temporal evolution with high teacher-student similarity.} Red indicates high similarity values. Best viewed in color.}
    \label{fig:dit}
\end{figure}

Figure \ref{fig:dit} illustrates the dynamic consensus region, which is initially concentrated around the target object (t $<$ 400), progressively diffusing to multi-scale regions (400 $\leq$ t $<$ 600), and stabilizing globally (t $\geq$ 600). Simply, we employ the consensus loss when t $<$ 400 during the training process, achieving promising results. We present the results for Inception Score (IS) \cite{salimans2016improved} and Fréchet Inception Distance (FID) \cite{heusel2017gans} for each method. Table \ref{tab:step KD} presents the results of step distillation experiments conducted on the CIFAR-10 dataset under the settings described in \cite{sun2023accelerating}. When distilled to 1-step sampling, the FID score decreased by 2. Table \ref{tab:diffusion model KD} illustrates the model distillation experiments on the ImageNet dataset. We utilize the DiT-base\cite{peebles2023scalable} model as the teacher to distill the DiT-small\cite{peebles2023scalable} student model. We use MSE loss as our baseline. The FID score decreased by 1 and the IS increased by 3 after training for 400k steps.

\begin{table}[h]
\renewcommand\arraystretch{1.3}
\setlength\tabcolsep{0.6mm}
\centering
\small
\footnotesize
\resizebox{0.38\textwidth}{!}{%
\begin{tabular}{c|c|c|c}
\begin{tabular}[c]{@{}l@{}}Sampling Steps\end{tabular} & Method & FID ($\downarrow$) & IS ($\uparrow$) \\ \Xhline{3\arrayrulewidth}
1                                                        & PD     &   \hspace{10pt} 15.926  \hspace{10pt}  &  \hspace{10pt} 7.764 \hspace{10pt}  \\
                                                         & PD+ADM &   \textbf{13.939}    &   \textbf{8.037}   \\ \hline
2                                                        & PD     &   7.322    &   8.653   \\
                                                         & PD+ADM &   \textbf{6.830}    &   \textbf{8.760}   \\ \hline
4                                                        & PD     &   4.667    &   9.020    \\
                                                         & PD+ADM &   \textbf{4.555}    &   \textbf{9.042}
\end{tabular}
}
\caption{\textbf{Results on Diffusion Step Distilltion.}}
\label{tab:step KD}
\vspace{-10pt}
\end{table}

\begin{table}[h]
\renewcommand\arraystretch{1.3}
\setlength\tabcolsep{0.6mm}
\centering
\small
\footnotesize
\resizebox{0.48\textwidth}{!}{%
\begin{tabular}{c|c|cc|cc|cc|cc}

\multirow{2}{*}{\begin{tabular}[c]{@{}c@{}}Model/\\ Method\end{tabular}}             & \multirow{2}{*}{CFG=4.0} & \multicolumn{2}{c|}{400K}        & \multicolumn{2}{c|}{300K}        & \multicolumn{2}{c|}{200K}       & \multicolumn{2}{c}{100K}        \\ \cline{3-10} 
                                                                                     &                          & FID-50K        & IS              & FID-50K        & IS              & FID-50K        & IS             & FID-50K        & IS             \\ \hline
\multirow{2}{*}{\begin{tabular}[c]{@{}c@{}}DiT-B/2\\ (teacher vanilla)\end{tabular}} &         $\checkmark$                 & 11.82          & 219.97          & 12.03          & 188.92          & 13.67          & 143.42         & 22.64          & 74.57          \\
                                                                                     &                          & 42.78          & 33.50           & 47.50          & 29.43           & 55.29          & 24.61          & 72.21          & 17.70          \\ \hline
\multirow{2}{*}{\begin{tabular}[c]{@{}c@{}}DiT-S/2\\ (student vanilla)\end{tabular}} &            $\checkmark$              & 14.83          & 123.75          & 17.05          & 105.53          & 21.94          & 79.20          & 36.13          & 43.62          \\
                                                                                     &                          & 67.18          & 20.30           & 71.42          & 18.91           & 78.58          & 16.64          & 94.00          & 13.16          \\ \hline
\multirow{2}{*}{DiT-S/2-MSE}                                                         &             $\checkmark$             & 13.67          & 136.73          & 15.75          & 112.74          & 18.93          & 93.63          & 31.98          & 48.85          \\
                                                                                     &                          & 62.08          & 22.27           & 65.75          & 20.77           & 72.71          & 18.30          & 86.43          & 14.52          \\ \hline
\multirow{2}{*}{DiT-S/2-ADM}                                                         &            $\checkmark$              & \textbf{12.75} & \textbf{139.41} & \textbf{14.77} & \textbf{124.65} & \textbf{17.95} & \textbf{99.08} & \textbf{30.04} & \textbf{53.98} \\
                                                                                     &                          & \textbf{61.04} & \textbf{22.76}  & \textbf{64.03} & \textbf{21.35}  & \textbf{70.60} & \textbf{18.83} & \textbf{85.22} & \textbf{14.75}

\end{tabular}
}
\caption{\textbf{Results on Diffusion Model Distilltion.}}
\label{tab:diffusion model KD}
\vspace{-10pt}
\end{table}



\begin{table*}[!h]
\renewcommand\arraystretch{1.3}
\setlength\tabcolsep{2.0mm}
\centering
\small

\footnotesize
\resizebox{0.75\textwidth}{!}{%
\begin{tabular}{c|cc|ccc}
\multirow{3}{*}{Method} & \multicolumn{2}{c|}{Same architecture style} & \multicolumn{3}{c}{Different architecture style}  \\ \cline{2-6} 
                        & ResNet-32$\times$4           & ResNet-56            & WRN-16-2        & ResNet-32$\times$4    & ResNet-50      \\
                        & ResNet-8$\times$4            & ResNet-20            & 0.5MobileNet-V2 & ShuffleNet-V2  & MobileNet-V2   \\ \Xhline{3\arrayrulewidth}
Teacher                 & 79.42                 & 72.34                & 73.26           & 79.42          & 79.34          \\
Student                 & 72.50                 & 69.06                & 64.60           & 71.82          & 64.60          \\ \hline
KD                      & 73.75                 & 71.05                & 68.91           & 75.35          & 68.42          \\
KD+ADM                   & \textbf{74.52}       & \textbf{71.46}       & \textbf{70.24}  & \textbf{75.73}  & \textbf{69.04}  \\
DKD                    & 76.24                 & 71.52                & 69.03           & 77.07          & 70.35          \\
DKD+ADM                  & \textbf{76.66}       & \textbf{71.58}      & \textbf{69.68} & 77.03          & \textbf{70.64} \\   
\end{tabular}
}
\caption{\textbf{Evaluation results in offline KD on CIFAR-100 dataset.} The upper and lower models denote teacher and student, respectively. }
\label{tab:cifar100-offlineKD}
\end{table*}

\begin{table*}[!h]
\renewcommand\arraystretch{1.3}
\setlength\tabcolsep{0.6mm}
\centering
\small
\footnotesize
\resizebox{0.85\textwidth}{!}{%
\begin{tabular}{c|cc|cccccccc}
 & Teacher & Student & KD    & KD+ADM     & DKD & DKD+ADM   & DisWOT & DisWOT+ADM  & NORM    & NORM+ADM  \\ \Xhline{3\arrayrulewidth}
top-1 & 73.31   & 69.75   & 71.03 & \textbf{71.25}  & 71.70 & \textbf{71.84} & 71.36 & \textbf{71.49}  & 72.03 & \textbf{72.15} \\
top-5 & 91.42   & 89.07   & 90.05 & 90.12 & 90.41 & 90.36 & 90.23 & 90.23 & 90.41 & \textbf{90.54} \\
\end{tabular}
}
\caption{\textbf{Top-1 accuracy (\%) in offline KD on the ImageNet validation.} We set ResNet-34 as the teacher and ResNet-18 as the student. All results are the average over 3 trials.}
\label{tab:imagenet-offlineKD}
\end{table*}

\begin{table*}[h!]
\center
\begin{small}
\resizebox{0.9\textwidth}{!}{%
\begin{tabular}{c|ccccccccccccc}
    $\alpha$ &  0.0 & 0.1  & 0.1 & 0.2 & 0.2   & 0.2   & 0.3 & 0.3 & 0.3  & 0.5 & 0.6 & 0.7 & 1.0   \\ \hline
    $\beta$  &  0.0 &  0.5  & 0.7 & 0.5 & 0.6   & 0.8   & 0.5 & 0.6 & 0.7  & 0.5 & 0.2 & 0.3 & 1.0   \\ \Xhline{3\arrayrulewidth}
    $stu\_acc$  & 71.12 & 71.42 & 71.37 & \textbf{71.48} & \textbf{71.59} & \textbf{71.48} & \textbf{71.69} & \textbf{71.60} & 71.37 & \textbf{71.52} & \textbf{71.55} & 71.43 & 71.31 \\
    $tea\_acc$  & 73.78 & 73.92 & 73.90 & 73.88 & \textbf{73.93} & 73.77 & \textbf{74.11} & \textbf{74.13} & \textbf{74.07} & \textbf{74.05} & \textbf{74.20} & \textbf{74.12} & \textbf{74.10} \\
\end{tabular}
}
\end{small}
\caption{\textbf{Different $\alpha$ and $\beta$.}}
\label{tab:alpha_beta_weight}
\vspace{-10pt}
\end{table*}

\vspace{-10pt}
\subsection{Extension to Offline KD}

ADM contains a teacher-discrepancy feature enhancement module and a student-similar feature knowledge distillation module. Evidently, the latter part is suitable for application in offline KD. To further validate the ADM, we also compare ADM with offline knowledge distillation approaches including KD and DKD that require a fixed and pre-trained teacher. Specifically, we set $\alpha$, $\beta$, $\gamma$ in Eq.(\ref{eq: decompose CAL}) and Eq.(\ref{eq:total loss}) as 1.0/0.0/0.0 and 0.1/0.0/0.0 for CIFAR-100 and ImageNet Dataset. Table \ref{tab:cifar100-offlineKD} reveals that ADM achieves superior performance over offline fashions. For example, incorporating ADM based on KD resulted in a 1.33\% improvement, and when applied to one of the state-of-the-art DKD methods, there was still an approximate increase of 0.3\% on CIFAR-100 dataset. Table \ref{tab:imagenet-offlineKD} shows that ADM increases KD accuracy by 0.22\% and NORM 0.12\% on ImageNet dataset.

\subsection{Extension to Multiple Teachers and Students.} 

Similar to DML and SwitOKD, ADM can be extended for multiple networks easily. As shown in Table \ref{tab:multiTS}, We adopt the same three-network settings as SwitOKD. 1T2S means that one teacher model and two student models, and 2T1S means that two teacher models and one student model. We achieve a higher accuracy student across all methods in the 2T1S.

\begin{table}[H]
\renewcommand\arraystretch{1.3}
\setlength\tabcolsep{0.6mm}
\centering
\small
\footnotesize
\resizebox{0.45\textwidth}{!}{%
\begin{tabular}{c|cccccc}
Backbone        & Vanilla & DML  & \begin{tabular}[c]{@{}c@{}}SwitOKD\\ (1T2S)\end{tabular} & \begin{tabular}[c]{@{}c@{}}SwitOKD\\ (2T1S)\end{tabular} & \begin{tabular}[c]{@{}c@{}}DML+ADM \\ (1T2S)\end{tabular} & \begin{tabular}[c]{@{}c@{}}DML+ADM \\ (2T1S)\end{tabular} \\ \Xhline{3\arrayrulewidth}
MobileNet-V2(S) & 66.01   & 72.28     & 72.20                                                    & 71.97                                                    & 71.69                                                & \textbf{72.51}                                       \\
WRN-16-2(S/T)   & 72.79   & 74.78     & 74.91                                                    & \textbf{75.07}                                                    & 74.80                                                & 74.09                                                \\
WRN-16-10(T)    & 79.30   & 80.32     & 80.55                                                    & 80.40                                                    & 80.58                                                & \textbf{80.72}                                               
\end{tabular}
}
\caption{\textbf{Accuracy (\%) comparison with 3 networks on CIFAR-100.} WRN-16-2 serves as either teacher (T) or student (S) for DML, while is treated as S for 1T2S and T for 2T1S.}
\label{tab:multiTS}
\vspace{-15pt}
\end{table}

\subsection{Ablation Studies}
\label{ablation studies}

\textbf{Hyper-Parameters Tuning.} We explore the impact of three different hyperparameter values on the performance. Table \ref{tab:alpha_beta_weight} and  \ref{tab:gamma_weight} report the student and teacher accuracy with different $\alpha$, $\beta$ and different $\gamma$ in ImageNet. We use ResNet-34 and ResNet-18 as teacher and student models, respectively. Experiments demonstrate that the same hyperparameters can be employed for distillation across distinct model pairs, and the configuration of hyperparameters has little influence on the results. For $\gamma$, due to the magnitude of $L_{feat}$, it is generally set lower than 0.1. For $\alpha$ and $\beta$, the results indicate that both parameters demonstrate stable increases when set around 0.5. It is worth noting that we observe a higher accuracy when the weights of $\alpha$ and $\beta$ are set close to the mIoU of foreground objects, as obtained in Figure \ref{fig:CAM_vis}. This finding substantiates that ADM method is indeed learning in an anticipated manner.

\begin{table}[h]
\vspace{-5pt}
\center
\begin{small}
\begin{tabular}{c|cc}
    $\gamma$     & 0.01  & 0.1   \\ \Xhline{3\arrayrulewidth}
    $stu\_acc$ & \textbf{71.59} & 71.48 \\
    $tea\_acc$ & \textbf{73.93} & 73.91
\end{tabular}
\end{small}
\caption{\textbf{Different $\gamma$.} $stu\_acc$ represents student's accuracy and $tea\_acc$ represents teacher's accuracy.} 
\label{tab:gamma_weight}
\vspace{-5pt}
\end{table}

\textbf{Effects of Feature Distillation Loss and ADM Loss.} This paper proposes three types of losses: $\mathcal{L}_{feat}$, $\mathcal{L}_{co}$ and $\mathcal{L}_{di}$. To validate the effectiveness of each loss, we conduct experiments to train students with these losses separately. The results on Table \ref{tab:different_loss} verify that both $\mathcal{L}_{feat}$ and $\mathcal{L}_{adm}$ can outperform the DML, also, the performance could be further boosted by combining them together. It is noteworthy that $L_{di}$ loss effectively enhances the performance of the teacher model. We also investigate the individual effectiveness of ADM in Table \ref{tab:only_CAL}.

\begin{table}[H]
\vspace{-5pt}
\center
\begin{small}
\resizebox{0.35\textwidth}{!}{%
\begin{tabular}{ccc|cc}
    $\mathcal{L}_{feat}$                  & $\mathcal{L}_{co}$            & $\mathcal{L}_{di}$            & $stu$\_$acc$ & $tea\_acc$ \\ \Xhline{3\arrayrulewidth}
    $ $ &                                                           &                               &     71.12      &      73.78      \\
    $\checkmark$ &                                                           &                               &     71.43      &      73.91      \\
                                                                 & $\checkmark$ &                               &    71.35     &     73.78    \\
                                                                 &                               & $\checkmark$ &    71.46     &     74.07    \\
                                                                 & $\checkmark$ & $\checkmark$ &     71.50    & 73.91        \\
                                                $\checkmark$     & $\checkmark$ & $\checkmark$ &     71.59    & 73.93        \\
                                                                 
\end{tabular} 
}
\end{small}
\caption{\small{\textbf{Effects of feature distillation loss and ADM loss.}}}
\label{tab:different_loss}
\vspace{-10pt}
\end{table}

\begin{table}[H]
\vspace{-10pt}
\center

\begin{small}
\resizebox{0.4\textwidth}{!}{%

\begin{tabular}{c|ccc}
    Backbone     & Vanilla  & DML  & ADM  \\ \Xhline{3\arrayrulewidth}
    S:ResNet-18	 & 70.16	& 71.12	& \textbf{71.33}  \\
    T:ResNet-34	 & 73.93	& 73.78	& 73.97  \\
    S:0.5MobileNetV2	 & 62.49	& 64.21	& \textbf{64.59}   \\
    T:ResNet-18	 & 70.16	& 69.00	& \textbf{70.58}  \\
    
\end{tabular}
}
\end{small}
\caption{\textbf{The individual effectiveness of ADM.} S: Student, T: Teacher.} 
\label{tab:only_CAL}
\vspace{-10pt}
\end{table}

\textbf{Different Types of ADM loss.} \label{sec: different cal loss} In this paper, we use cross-entropy(CE) to strengthen the ADM loss for brevity. Table \ref{tab:different_cal_loss} shows different forms of ADM loss. Adopting CE and KD loss helps us to achieve better performance.

\begin{table}[H]
\center
\begin{small}
\begin{tabular}{ll|ll}
     $\mathcal{L}_{co}$ & $\mathcal{L}_{di}$  & $stu\_acc$ & $tea\_acc$ \\ \Xhline{3\arrayrulewidth}
    $\mathcal{L}_{ce}$ &  $\mathcal{L}_{ce}$ &     71.41    &     74.03     \\
    $\mathcal{L}_{kd}$ &  $\mathcal{L}_{ce}$ &   71.59     &    73.93     \\
    $\mathcal{L}_{kd}$ &  $\mathcal{L}_{kd}$ &  71.45      &     73.98    
\end{tabular} 
\end{small}
\caption{\small{\textbf{Different types of ADM loss.}}}
    \label{tab:different_cal_loss}
\vspace{-5pt} 
\end{table}

\textbf{Combined with More Methods.} ADM is a plug-and-play module that can effectively enhance the performance of existing methods. In this paper, when combined with DML, ADM can basically achieve sota in all settings, and when combined with stronger methods, better results can be achieved. As shown in Table \ref{tab:imagenet}, KDCL achieves the sota performance in some settings, and Table \ref{tab:com_with_KDCL} shows several groups of comparative experiments combined with KDCL and SHAKE on the ImageNet dataset.

\begin{table}[H]
\vspace{-5pt}
\center
\begin{small}
\resizebox{0.45\textwidth}{!}{%
\begin{tabular}{c|ccccc}
    Backbone     & Vanilla  & KDCL  & KDCL+ADM   & SHAKE & SHAKE+ADM \\ \Xhline{3\arrayrulewidth}
    S:ResNet-18	 & 70.16	& 71.54	& \textbf{71.78} (+0.24) & 71.65 & \textbf{71.75} (+0.10)\\
    T:ResNet-34	 & 73.93	& 74.36	& 74.34  & 73.45 & 73.50\\
    
\end{tabular}
}
\end{small}
\caption{\textbf{Combined with KDCL.} S: Student, T: Teacher. } 
\label{tab:com_with_KDCL}
\vspace{-15pt}
\end{table}

\begin{figure}[H]
    \centering
    \vspace{-10pt}
    \subfigure[t-SNE]{\includegraphics[scale=0.148]{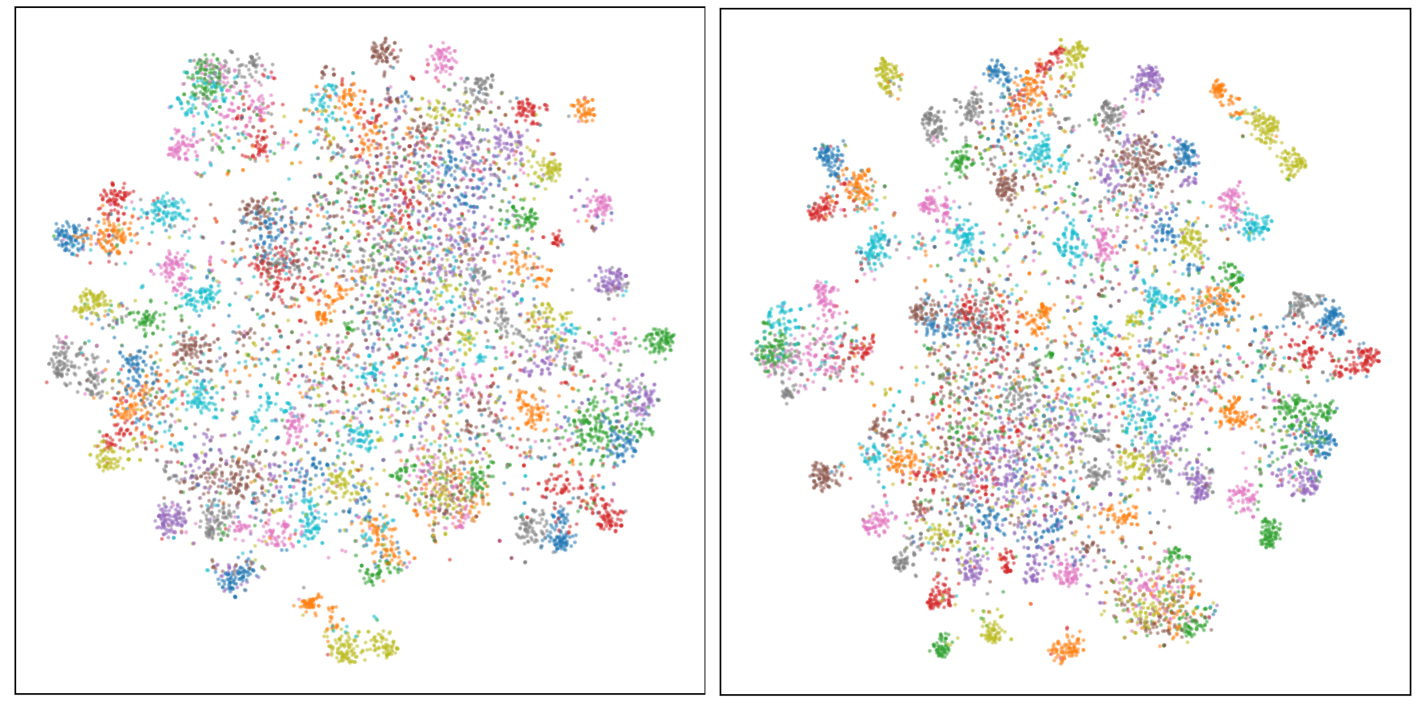}}
    \subfigure[Correlation matrix]{\includegraphics[scale=0.185]{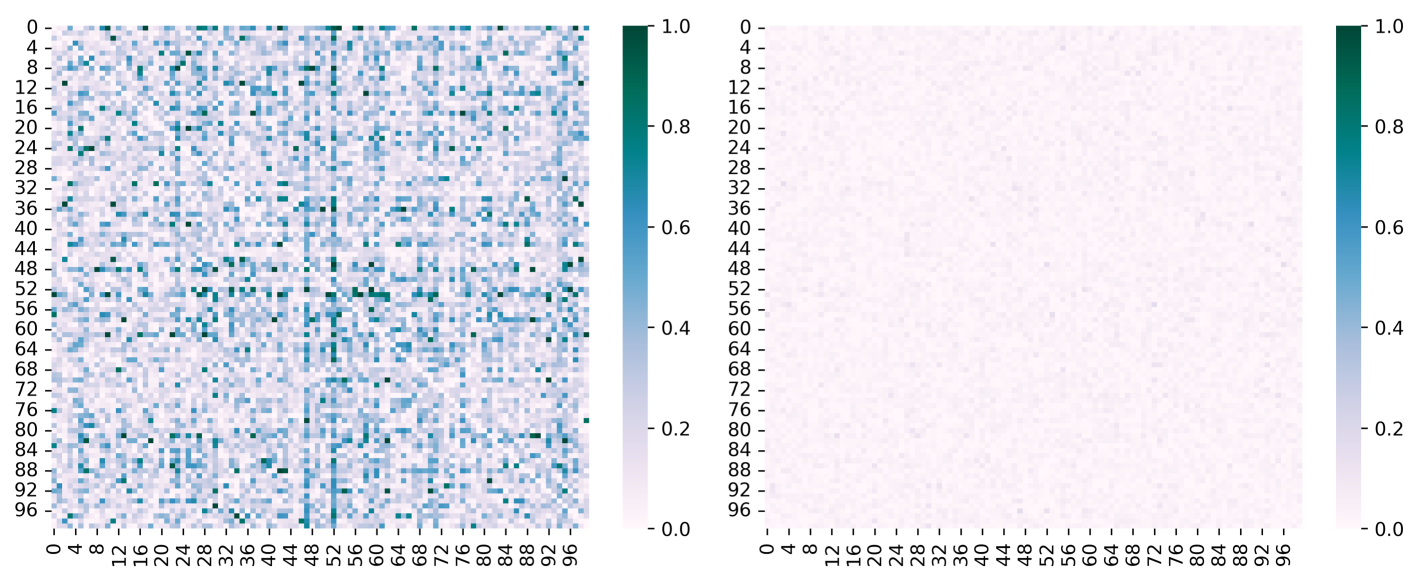}}

    \caption{\textbf{Visualzaiton}. Obviously, ADM (right) results in more discriminative features, reduced prediction discrepancies with teachers, and enhanced calibration than DML (left). }
    \label{fig:visualization}
\vspace{-10pt}
\end{figure}

\textbf{Visualization.} In addition, we present visualizations from three perspectives in Figure \ref{fig:visualization} (with setting teacher as ResNet32$\times$4 and student as ResNet8$\times$4 on CIFAR-100).

\section{Conclusion}
\label{sec:conclusion}
In this paper, we present Asymmetric Decision-Making (ADM), an innovative online knowledge distillation method that unifies consensus and divergence feature learning and can seamlessly integrates with existing methods to leverage feature richness. Based on our insight into feature similarity, ADM employs a Consensus Learning on the student model to facilitate the learning of fundamental spatial knowledge, and a Divergence Learning on the teacher model to continually enhance its capabilities and drive the student model to learn and catch up. This dynamic interaction between the teacher and student models epitomizes the principle that "iron sharpens iron," thereby achieving superior performance in online knowledge distillation. Extensive experiments show our superiority in various benchmark tasks. We hope this paper will provide some insights for future OKD research.

\nocite{langley00}

\bibliography{example_paper}
\bibliographystyle{icml2025}

\newpage
\appendix
\onecolumn

\section{Implementation of ADM} \label{sec:impl}
This section presents the implementation code of ADM in PyTorch, as shown in Algorithm \ref{algo:ctc}. With the intermediate features of student and teacher models, our ADM leverages quite simple inputs and computations and is easy to implement. The total loss is the sum of DML loss and ADM loss.

\begin{algorithm*}[htp] \tiny\tiny
    \caption{Implementation code of ADM in a PyTorch-like style.}
    \label{algo:ctc}
    \footnotesize
    \begin{alltt}
    \color{ForestGreen}
# tea, stu: teacher and student Module
# alpha, beta, gamma: hyper-parameters
# feat_tea, feat_stu: intermediate features of teacher and student
\end{alltt}

\begin{alltt}
{\color{ForestGreen}import} {\color{blue}{torch.nn}} {\color{ForestGreen}as} {\color{blue}{nn}}
{\color{ForestGreen}import} {\color{blue}torch.nn.functional} {\color{ForestGreen}as} {\color{blue}{F}}

\begin{minipage}{0.5\textwidth} 
{\color{ForestGreen}class} {\color{blue}ADM}(nn.Module):
    {\color{ForestGreen}def} {\color{blue}__init__}({\color{ForestGreen}self}, tea, stu, alpha, beta, gamma):
        {\color{ForestGreen}super}({\color{blue}ADM}, {\color{ForestGreen}self}).{\color{blue}__init__}()
        {\color{ForestGreen}self}.tea = tea
        {\color{ForestGreen}self}.stu = stu
        {\color{ForestGreen}self}.alpha = alpha
        {\color{ForestGreen}self}.beta = beta
        {\color{ForestGreen}self}.gamma = gamma

    {\color{ForestGreen}def} {\color{blue}forward}({\color{ForestGreen}self}, feat_stu, feat_tea, target):
        feat_loss = {\color{ForestGreen}self}.feat_loss(feat_stu, feat_tea, tea_detach=True)
        adm_loss = {\color{ForestGreen}self}.adm_loss(feat_stu[-1], feat_tea[-1], target)

        {\color{ForestGreen}return} adm_loss + {\color{ForestGreen}self}.gamma * feat_loss

    {\color{ForestGreen}def} {\color{blue}adm_loss}({\color{ForestGreen}self}, stufm, teafm, target):
        sim = F.cosine_similarity(stufm, teafm, dim=1).detach()
        attn_adv = (1 - sim) / (
            (1 - sim).mean(dim=(2, 1), keepdim=True) + 1e-5
        ).unsqueeze(1)
        attn_co = (1 + sim) / (
            1 + sim.mean(dim=(2, 1), keepdim=True) + 1e-5
        ).unsqueeze(1)
        logits_tea = {\color{ForestGreen}self}.tea.fc(
            {\color{ForestGreen}self}.tea.avgpool(
                attn_adv * F.relu(teafm)).view(attn_adv.size(0), -1)
        )
        logits_stu = {\color{ForestGreen}self}.stu.fc(
            {\color{ForestGreen}self}.stu.avgpool(
                attn_co * F.relu(stufm)).view(attn_co.size(0), -1)
        )
        {\color{ForestGreen}return} {\color{ForestGreen}self}.alpha * F.cross_entropy(logits_stu, target) + \textbackslash
            {\color{ForestGreen}self}.beta * F.cross_entropy(logits_tea, target)

    {\color{ForestGreen}def} {\color{blue}feat_loss}({\color{ForestGreen}self}, feat_stu, feat_tea, tea_detach=True):
        loss = 0
        N = len(feat_stu)
        for idx in range(1, N):
            ft = feat_tea[idx]
            fs = feat_stu[idx]
            if tea_detach:
                ft = ft.detach()
            loss += F.mse_loss(fs, ft) / 2 ** (N - idx)
        {\color{ForestGreen}return} loss

\end{minipage}

\end{alltt}

\end{algorithm*}

\section{Properties of ADM} \label{sec:properties}

\begin{figure}[htbp]
    \centering
    \subfigure[student KL Divergence] 
    {\label{fig:prop-a}\includegraphics[width=0.4\linewidth,height=0.28\linewidth]{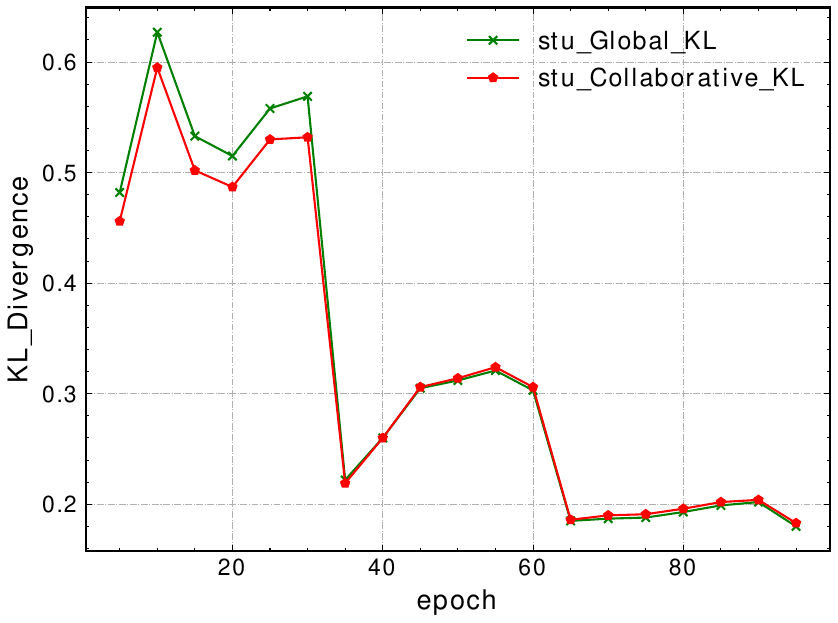}}
    \hspace{6mm}
    \subfigure[teacher CE Loss]
    {\label{fig:prop-b}\includegraphics[width=0.4\linewidth,height=0.28\linewidth]{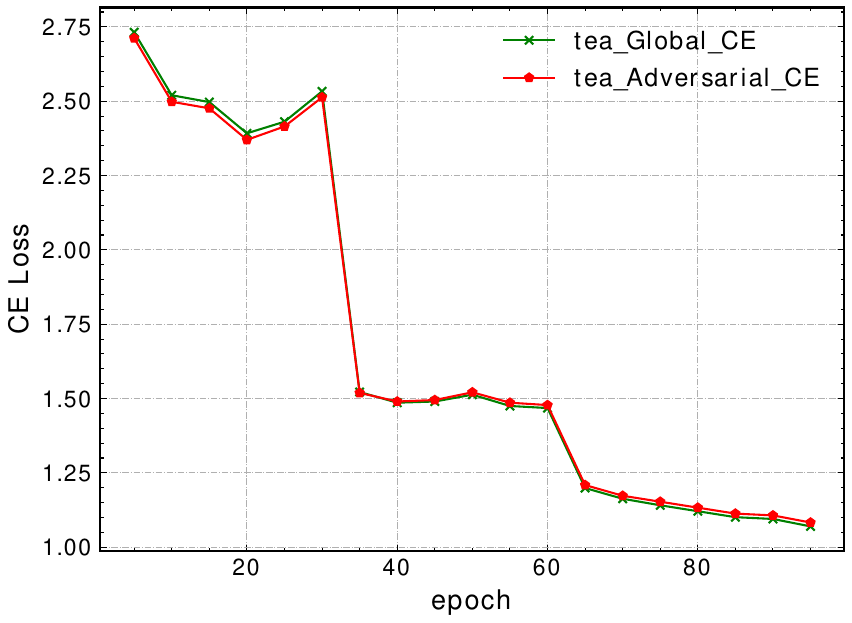}}
    \caption{\textbf{Illustration of the properties of ADM loss.}
    }
    \label{fig:prop}
\vspace{-10pt}
\end{figure}

\textbf{\emph{Consensus Learning enables student models to concentrate on relatively simple knowledge in the early training stage but hard knowledge in the last training stage, similar to Divergence Learning for teacher models.}} It is in line with the principles of Curriculum Learning. As summarized in Figure \ref{fig:prop-a}, we calculate the collaborative KL divergence between the logits of student and teacher models weighted by feature similarity on ImageNet with teacher ResNet-34 and student ResNet-18. Figure \ref{fig:prop-b} shows the difference between the adversarial CE loss weighted by similarity and the CE loss calculated after Global Average Pooling (GAP). We have discovered that ADM bridges the capacity gap between student and teacher models in the early training stage. However, in the last training stage, it is essential to enable both the student and teacher models to focus more on challenging knowledge. We also found that $L_{di}$ gives higher weights to some regions in the early stage, however, the weight distribution across various areas in the later stage becomes fairly uniform, resulting in a negligible level of noise as compared to the global average pooling. So it does not encourage the teacher model to do correct classification on the background information.

\begin{figure}[htbp]
    \centering
    \subfigure[CE Loss]
    {\label{fig:400-c}\includegraphics[width=0.45\linewidth,height=0.3\linewidth]{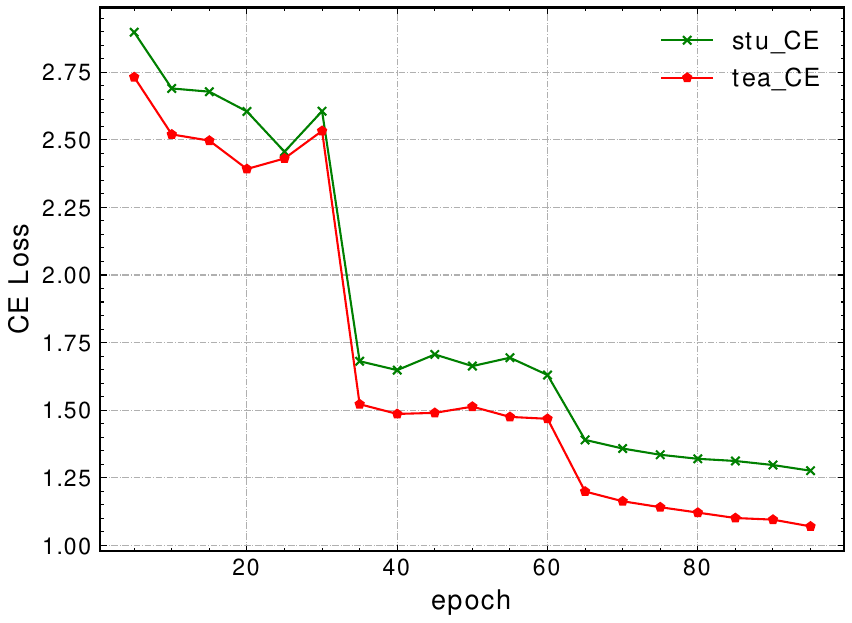}}
    \hspace{6mm}
    \subfigure[KL loss]
    {\label{fig:400-d}\includegraphics[width=0.45\linewidth,height=0.3\linewidth]{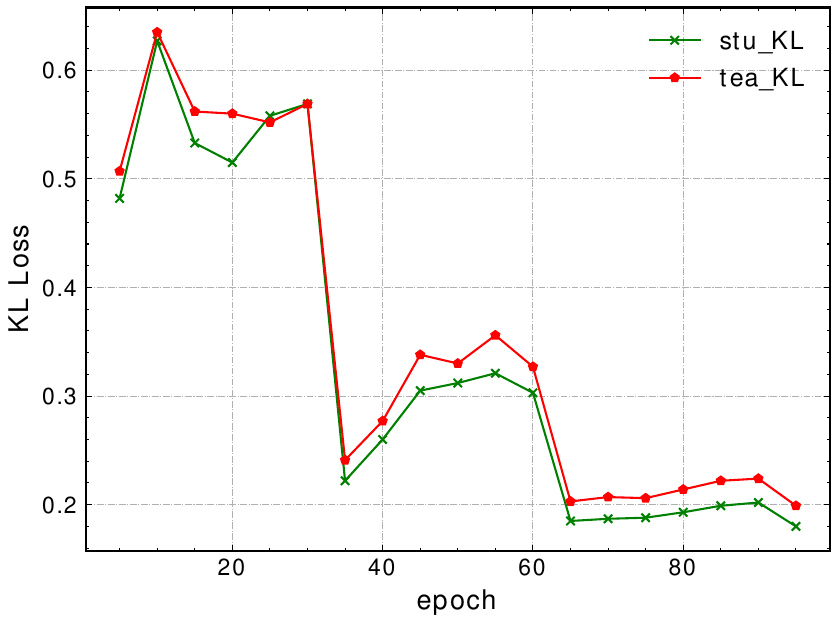}}
    \caption{\textbf{The CE and KL loss for both the teacher and student models during the training process.}
    }
    \label{fig:400}
\end{figure}

\textbf{\emph{Teacher models exhibit more confidence than students and are subject to more constraints imposed by the student models.}} Figure \ref{fig:400-c} and \ref{fig:400-d} present the CE loss and KL loss for both the teacher and student models during the training process. It is obvious that the teacher model exhibits more confidence in its predictions, yet it is also subject to greater distillation constraints imposed by the student model. Consequently, it is essential to employ divergence loss to enhance the advantages of the teacher model continuously.

\textbf{Interpreting $L_{di}$ from a Statistical Perspective.}

We discovered that primarily enables the teacher model to concentrate more on the areas surrounding foreground objects in the early training stage. In the later training stage, the background noise introduced by supervision is minimal. Based on the experimental setting of ResNet34-ResNet18 pairs on ImageNet, this observation was validated through the calculation of the similarity matrix $\frac{1-\mathcal{S}}{1-\Bar{\mathcal{S}}}$ in Eq.(8) during the training process. Table \ref{tab: similarity matrix value} shows the minimum, maximum, and variance values of the similarity matrix during the training process. We found that gives higher weights to some regions in the early stage, however, the weight distribution across various areas in the later stage becomes fairly uniform, resulting in a negligible level of noise introduced as compared to global average pooling. So it does not encourage the teacher model to do correct classification on the background information.

\begin{table*}[h]
    \center
    \begin{small}
    \begin{tabular}{c|ccccccccc}
    epoch	& 5	&10	&15	&20	&30	&40	&60	&80	&95  \\ \Xhline{3\arrayrulewidth}
    minimum	& 0.6768	& 0.6787	& 0.6911	& 0.6980	& 0.6870	& 0.8431	& 0.8581	& 0.8476	& 0.8441 \\
    maximum	& 1.3010	& 1.3097	& 1.2990	& 1.2875	& 1.1959	& 1.1213	& 1.1233	& 1.1370	& 1.1421 \\
    variance &	0.0301	& 0.0302	& 0.0272	& 0.0254	& 0.0213	& 0.0053	& 0.0047	& 0.0057	& 0.0059 \\
    \end{tabular} 
    \end{small}
    
    \caption{\small{\textbf{The value of similarity matrix during the training process.}}}
    \label{tab: similarity matrix value}
\end{table*}

\textbf{Why the performance of the teacher model with the method in this paper will decline in some cases?}
This is prevalent within current existing OKD approaches. The underlying cause is the inferior performance of the student model compared to the teacher model, leading to excessive constraints on the latter. Consequently, this impacts the teacher model's performance adversely. 

\section{Poor performance teacher distillation} \label{sec:poor}

\begin{figure}[h]
    \centering
    \subfigure[ResNet32$\times$4 $\rightarrow$ ResNet8$\times$4]
    {\includegraphics[width=0.38\linewidth,height=0.3\linewidth]{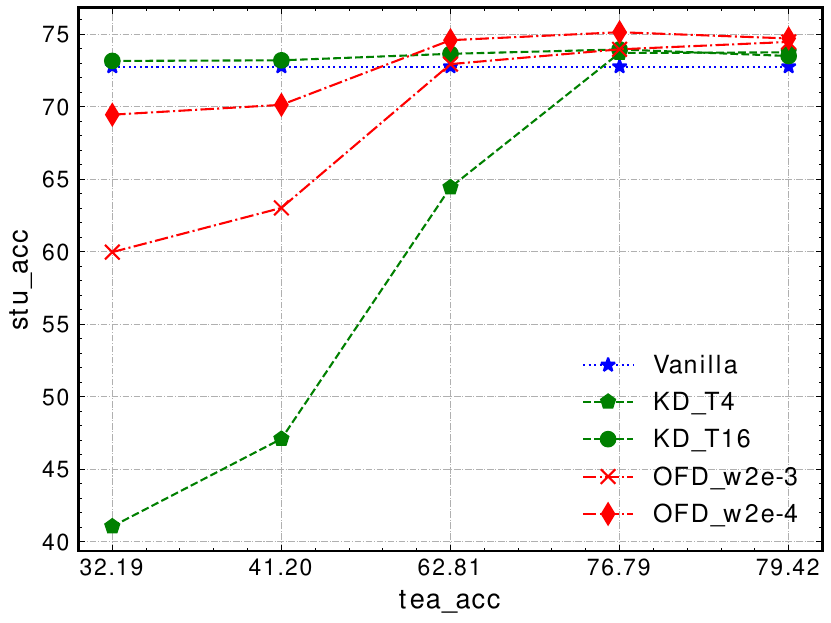}}
    \hspace{6mm}
    \subfigure[ResNet50 $\rightarrow$ MobileNetV2]
    {\includegraphics[width=0.38\linewidth,height=0.3\linewidth]{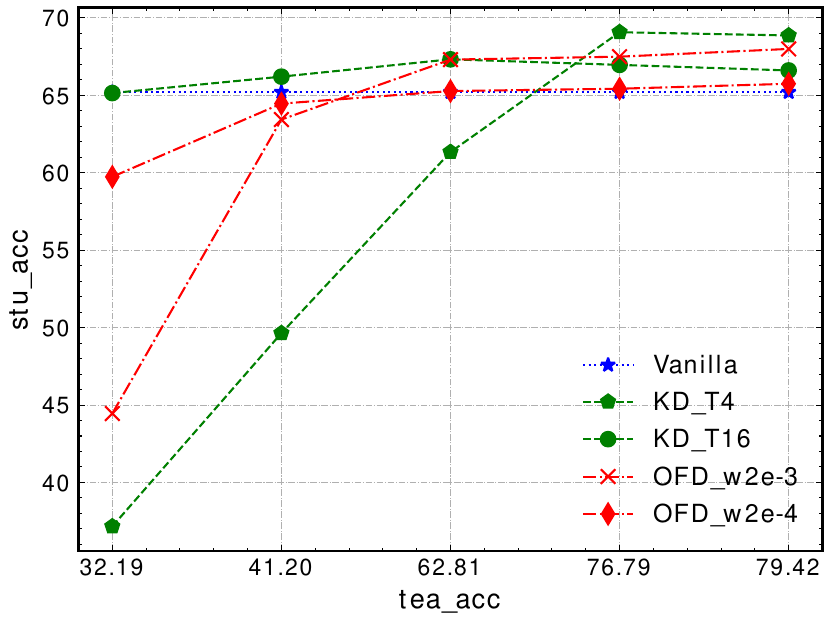}}
    \caption{\textbf{Comparisons of Logits-based KD and Feature-based KD(OFD) with different poorly-trained teacher model on CIFAR100.} We report mean accuracy over 5 runs. (a) The ResNet-8x4 students are trained using different temperatures (T) and different loss weights (w) with ResNet-32x4 teachers. (b) The MobileNet-V2 students are trained using different temperatures (T) and different loss weights (w) with ResNet-50 teachers.}
    \label{fig: pptd}
    \vspace{-10pt}
\end{figure}

As shown in Figure~\ref{fig: pptd}, we observe that a poorly-trained teacher with much lower accuracy can still improve the student significantly with appropriate temperate(T) in logits-based KD. But adjusting the weight of feature distillation loss in feature distillation does not achieve comparable results. Because the poor performance teacher provides an unclear optimization direction for the feature learning of the student network. In ablation studies, we investigated several strategies for filtering noise from features, with the anticipation that future work will delve further into this issue.

\begin{figure}[htbp]
    \centering

    \subfigure[ResNet-34 $\rightarrow$ ResNet-18]
    {\label{fig:1-c}\includegraphics[width=0.43\linewidth,height=0.3\linewidth]{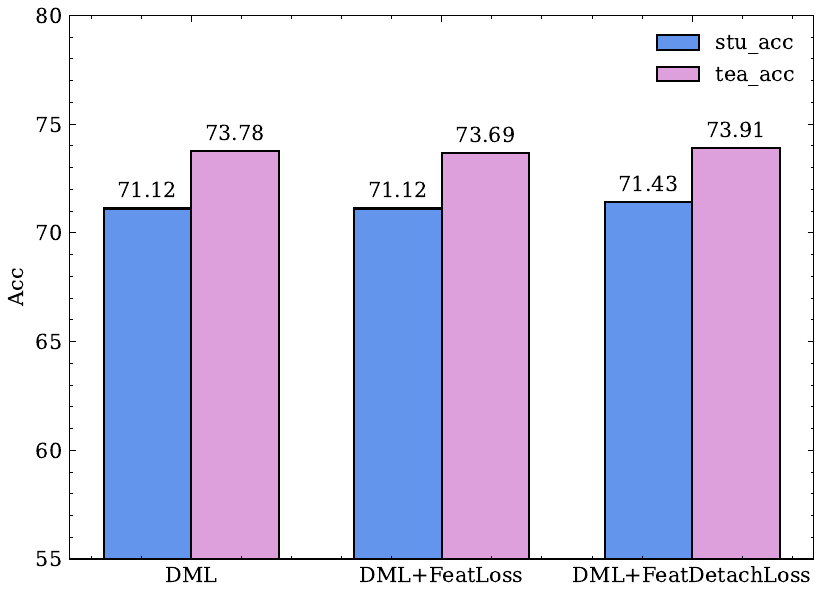}}
    \hspace{6mm}
    \subfigure[ResNet-18 $\rightarrow$ 0.5MobileNet-V2]
    {\label{fig:1-d}\includegraphics[width=0.43\linewidth,height=0.3\linewidth]{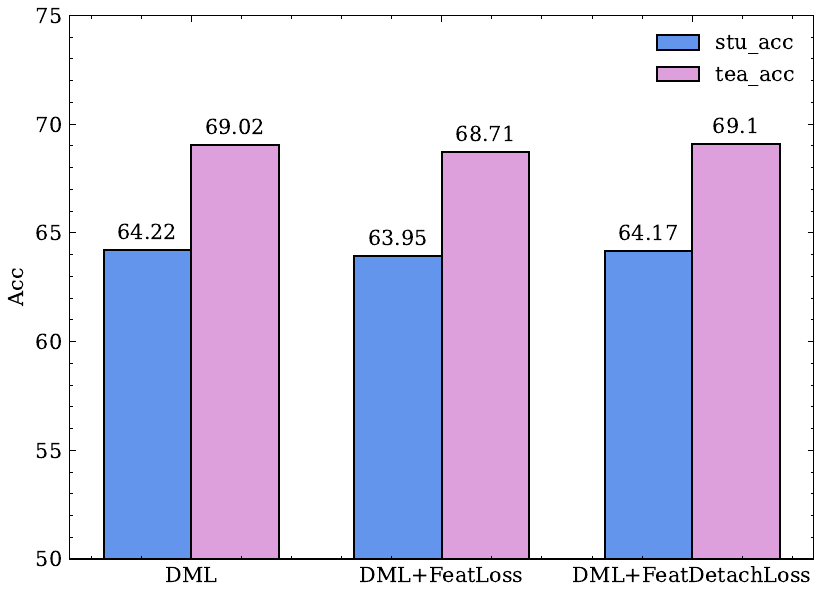}}
    \caption{\textbf{Illustration of the problems suffered by directly aligning the feature maps on OKD.} Within the DML framework on ImageNet, Feature Loss (FeatLoss) incorporates the Mean Squared Error (MSE) loss between the features of different layers from both teacher and student models. Feature Detach Loss (FeatDetachLoss) means that detaching the gradients associated with the teacher model mimicking the student's features.
}
    \label{fig:100}
\end{figure}

Despite the promising performance of offline feature distillation, as shown in Figure~\ref{fig:100}, we notice that \textbf{directly mimicking the student’s features could enforce overly strict constraints on the teacher}. This leads to sub-optimal results in teacher model, subsequently leading to deterioration in the performance of student model. We hypothesize that detaching the gradient of making teacher mimic student's feature for OKD helps to mitigate the overly strict restrictions problem when minimizing feature discrepancy between student and teacher.

\textbf{More strategies for feature distillation with poor-performance teacher.} \label{sec: feature distillation strategy} In the early training stage of OKD, the teacher model demonstrates poor performance, which is unfavorable for feature distillation. Therefore, we investigate several strategies to alleviate this problem, as shown in Table \ref{tab:different_feat_loss}. Specifically, $\mathcal{L}_{norm}$ performs L2-Norm along the dimensions of $h*w$, $\mathcal{L}_{relu}$ adds ReLU activation before MSE loss function, $\mathcal{L}_{drop}$ drops top 1/3 MSE loss as noise. This implies that filtering away noises from the features can lead to improved effects.

\begin{table}[h]
\center
\begin{small}
\begin{tabular}{l|ll}
       & $stu\_acc$ & $tea\_acc$ \\ \Xhline{3\arrayrulewidth}
    $\mathcal{L}_{norm}$ &      71.32    &     73.82     \\
    $\mathcal{L}_{relu}$ &     71.44     &    73.96     \\
    $\mathcal{L}_{drop}$ &    71.54      &     74.05   
\end{tabular} 
\end{small}
\caption{\small{\textbf{Different Feature loss.}}}
\label{tab:different_feat_loss}
\end{table}


\section{More Ablation Studies}

\textbf{Adjusting the distillation loss weight $\lambda$ imposed by the student on the teacher.} As shown in Table \ref{tab:different_lambda} We conducted additional experiments by merely adjusting the distillation loss weight imposed by the student on the teacher. Our findings indicate that decreasing the loss of weight in distillation on the teacher can lead to an enhanced performance improvement of the teacher model. However, we also adhere to the experimental settings employed by other OKD methods, utilizing the default weight parameters.

\begin{table}[!h]
\center
\begin{small}
\resizebox{0.45\textwidth}{!}{%
\begin{tabular}{c|ccccc}

Backbone	& Vanilla	& DML	& ADM	& ADM+DML(=1)	& ADM+DML(=0.3) \\ \Xhline{3\arrayrulewidth}
S:ResNet-18	& 70.16	& 71.12	& 71.33	& \textbf{71.59}	& 71.58 \\
T:ResNet-34	& 73.93	& 73.78	& 73.97	& 73.93	& \textbf{74.21} \\
\end{tabular} 
}
\end{small}
\caption{\small{\textbf{Adjusting the distillation loss weight $\lambda$ imposed by the student on the teacher.}}}
\label{tab:different_lambda}
\end{table}

\textbf{More feature-based OKD baselines comparison} The performance of existing feature-based OKD methods in various experiments was found to be suboptimal compared to alternative methods. Table \ref{tab:feature distill} presents the performance comparison between AFD \cite{chung2020feature} and AFID \cite{su2021attention} methods on ImageNet, as well as the performance when ADM is integrated with these two approaches using default hyper-parameters.

\begin{table}[h]
\center
\begin{small}
\resizebox{0.6\textwidth}{!}{%
\begin{tabular}{c|cccccc}

Backbone	& Vanilla	& AFD	& AFD+ADM	& AFID	& AFID+ADM	& ADM \\ \Xhline{3\arrayrulewidth}
S:ResNet-18	& 70.16	& 70.88	& 70.91	& 70.91	& 71.05(+0.14)	& 71.33 \\
T:ResNet-34	& 73.93	& 73.00	& 73.31(+0.31)	& 74.18	& 74.18	& 73.97 \\
\end{tabular} 
}
\end{small}
\caption{\small{\textbf{More feature-based OKD baselines comparison.}}}
\label{tab:feature distill}
\end{table}

\textbf{Experiments on noisy labels.} Specifically, we randomized the labels of 10\%, 30\%, and 50\% of the training data on CIFAR-100 and conducted comparative experiments under these noisy label conditions. Our method still achieved improvements, as shown in \ref{tab:noisy-label-rebuttal-1}. We attribute these gains to the collaborative loss, which likely enhances noise filtering through feature similarity, enabling the student model to leverage the teacher model for feature denoising and thereby yielding better performance. We run all methods five times with different seeds and obtain the average accuracy.

\begin{table}[h]
\renewcommand\arraystretch{1.3}
\setlength\tabcolsep{0.6mm}
\centering
\small
\footnotesize
\resizebox{0.5\textwidth}{!}{
\begin{tabular}{c|ccc|ccl|ccc}
                           & \multicolumn{3}{c|}{10\%} & \multicolumn{3}{c|}{30\%}                                              & \multicolumn{3}{c}{50\%} \\ \cline{2-10} 
\multirow{-2}{*}{Backbone} & Valinna  & DML    & ADM   & Valinna & DML   & \multicolumn{1}{c|}{ADM}                             & Valinna  & DML   & ADM   \\ \hline
ResNet8×4                  & 68.50    & 70.41  & \textbf{70.64} & 63.97   & 66.56 & \textbf{66.66} & 57.73    & 61.21 & \textbf{61.40} \\
ResNet32×4                 & 74.57    & 75.78  & \textbf{76.15} & 69.33   & 70.82 & \textbf{71.14} & 63.04    & 64.92 & \textbf{65.03} \\ \hline
0.5MobileNetV2             &  52.23        &    62.32    &    \textbf{63.59}   &  42.46       &   56.17    &   \textbf{57.74}    &       32.78   &    50.04   &    \textbf{51.77}   \\
WRN-16-2                   &  67.89        &   69.07   &    \textbf{70.22}     &     63.72    &   63.92    &                          \textbf{64.99}      &     58.29     &   58.20    &  \textbf{59.34}     

\end{tabular}
}

\caption{\textbf{Top-1 Accuracy comparison on CIFAR-100 on noisy label settings.} The upper and lower models denote student and teacher, respectively.}
\label{tab:noisy-label-rebuttal-1}

\end{table}

\section{Different forms of ADM} \label{sec:forms}

In order to prevent over-confidence in student and teacher models, we investigate the combination of CE loss with KD loss in ADM as the following, which achieved competitive results:

\vspace{-10pt}
\scriptsize
\begin{align}
     \label{eq: loss_co_kd}
     \mathcal{L}_{co}^{kd} = {\mathcal{L}_{kd}}\left(W^s\left(\psi_{avg}\left (\frac{1+\mathcal{S}}{1+\Bar{\mathcal{S}}} \cdot F^s\right)\right) , 
     W^t\left(\psi_{avg}\left (\frac{1+\mathcal{S}}{1+\Bar{\mathcal{S}}} \cdot F^t\right)\right) \right)
 \end{align}

\vspace{-10pt}
\small
\begin{align}
     \label{eq: loss_ad_kd}
     \mathcal{L}_{di}^{kd} = {\mathcal{L}_{kd}}\left(W^t\left(\psi_{avg}\left (\frac{1-\mathcal{S}}{1-\Bar{\mathcal{S}}} \cdot F^t\right)\right), 
     (1 - \delta) y + \delta * p^t\right)
 \end{align}

where, $\delta$ is a momentum parameter to control how much we trust the prediction from the teacher, y is the one-hot target, and $p^t$ is the teacher's prediction in the last epoch. In our experiments, we increase the value of $\delta$ linearly from 0.2 to 0.6.


\section{More Visualization} \label{sec:more vis}
\begin{figure}[h]
    \centering
    
    

    

        \includegraphics[width=\linewidth]{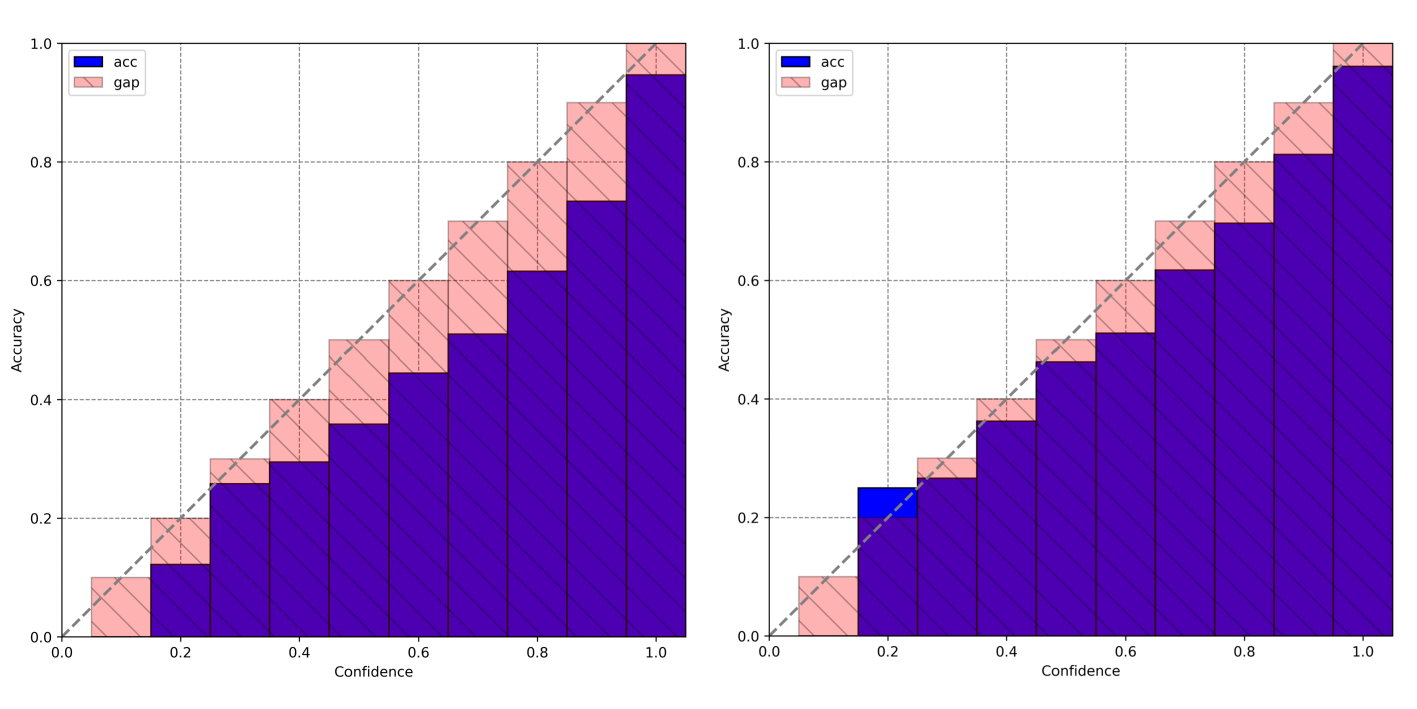}

    \caption{\textbf{Calibration Curves.} Obviously, ADM (right) results in more reduced prediction discrepancies with teachers, and enhanced calibration than DML (left). }
    \label{fig:more visualization}
    
\end{figure}

In addition, we present visualizations from additional perspectives in Figure \ref{fig:more visualization} (with setting teacher as ResNet32$\times$4 and student as ResNet8$\times$4 on CIFAR-100).

\section{Test Acc curves} \label{sec:acc curves}

\begin{figure*}[h]
    \centering
    \subfigure[stu test acc]
    {\includegraphics[width=0.3\linewidth,height=0.21\linewidth]{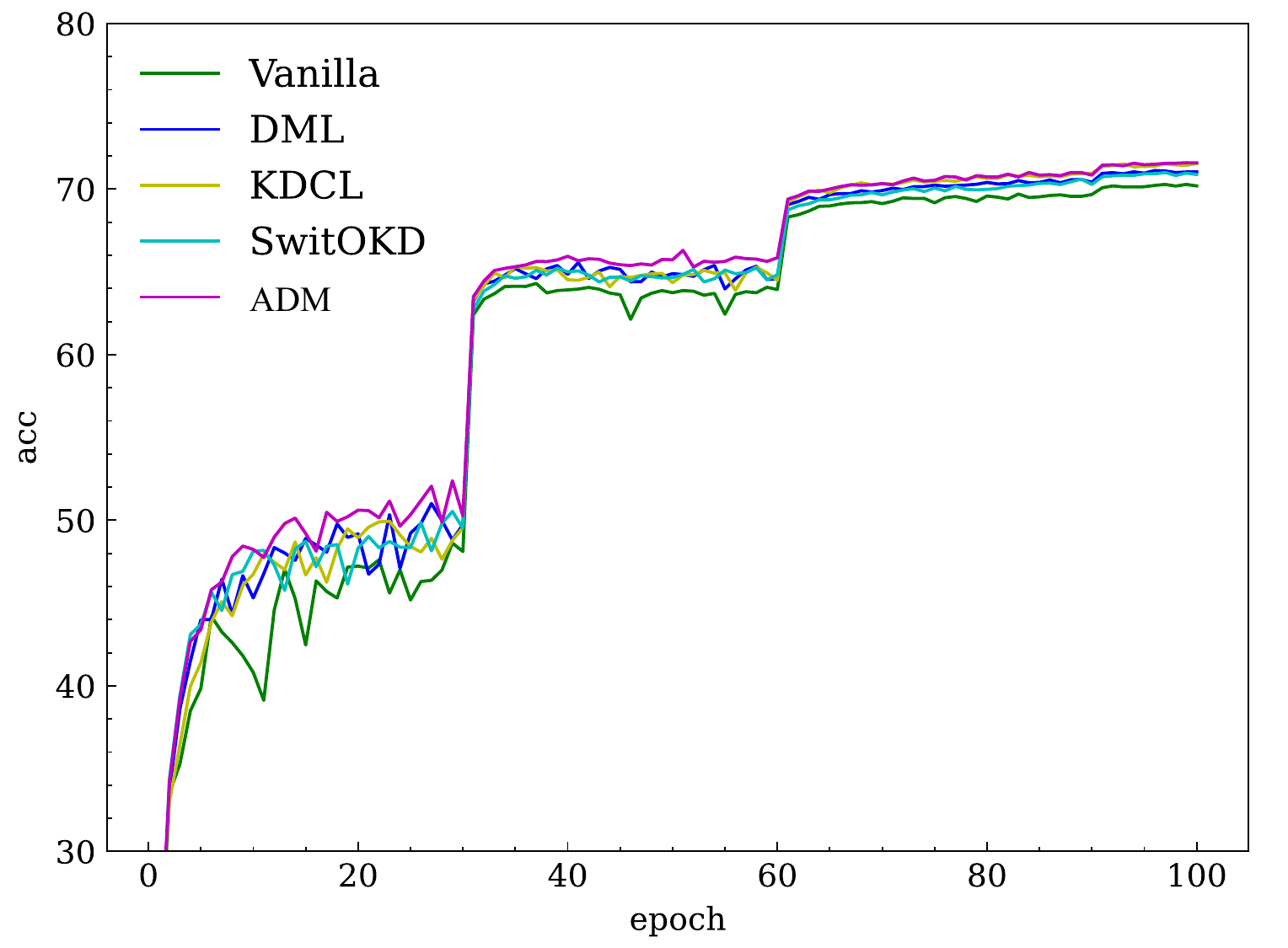}}
    \hspace{1mm}
    \subfigure[tea test acc]
    {\includegraphics[width=0.3\linewidth,height=0.21\linewidth]{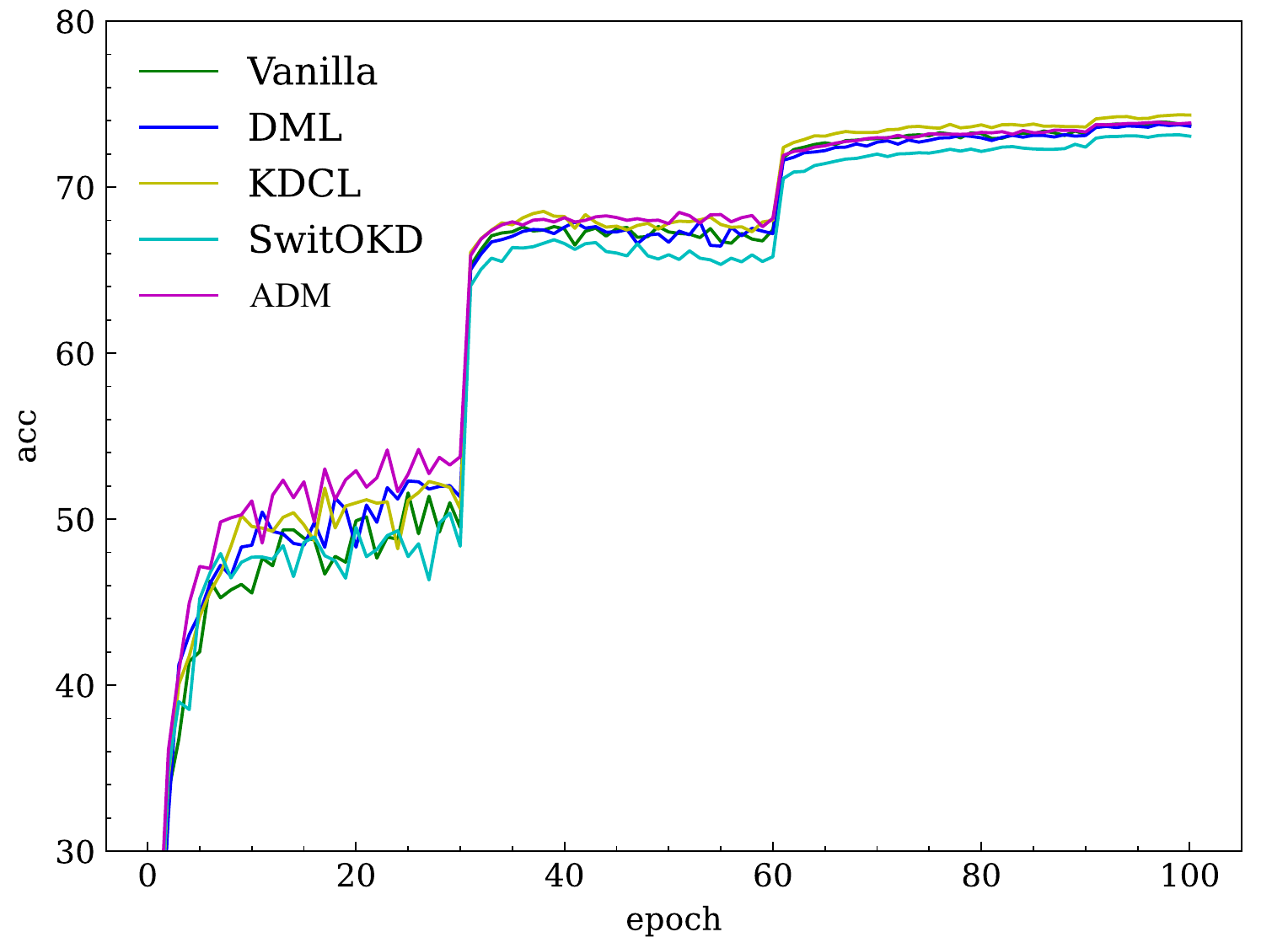}}
    \hspace{1mm}
    \subfigure[mean test acc]
    {\includegraphics[width=0.3\linewidth,height=0.21\linewidth]{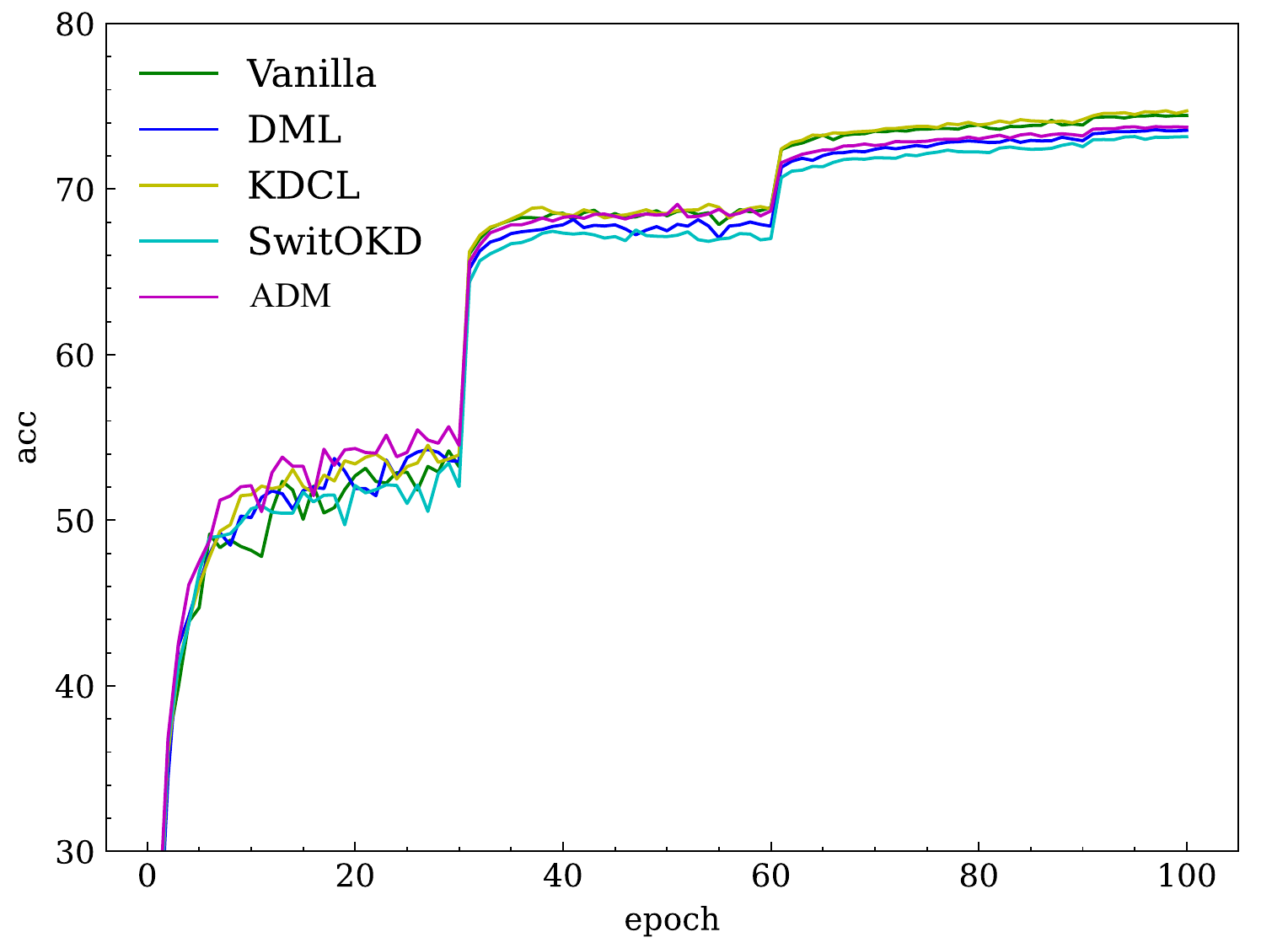}}
    \caption{\textbf{Testing Acc during training.} ResNet-34 $\rightarrow$ ResNet18 on ImageNet.
    }
    \label{fig:3}
    \subfigure[stu test acc]
    {\includegraphics[width=0.3\linewidth,height=0.21\linewidth]{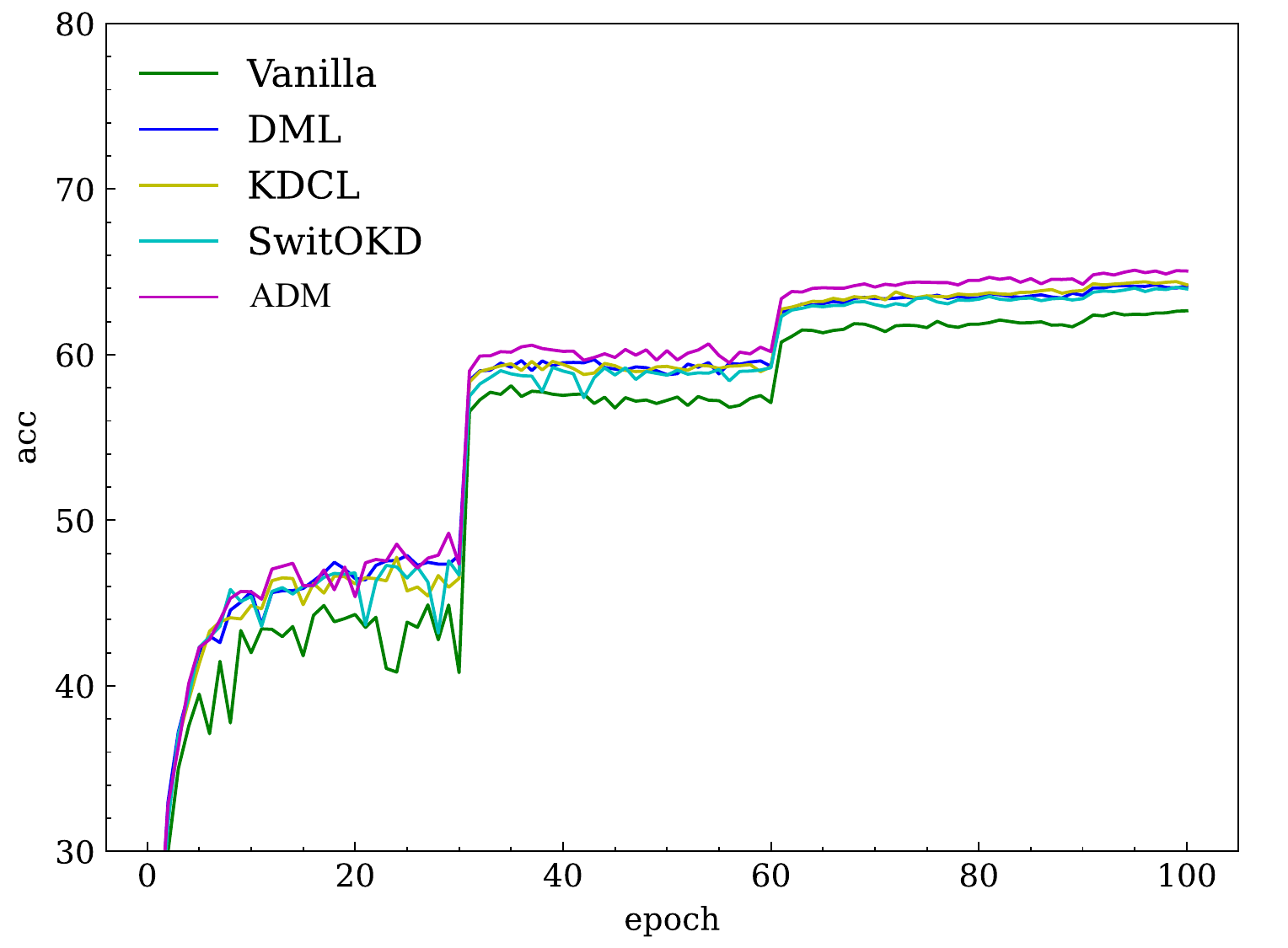}}
    \hspace{1mm}
    \subfigure[tea test acc]
    {\includegraphics[width=0.3\linewidth,height=0.21\linewidth]{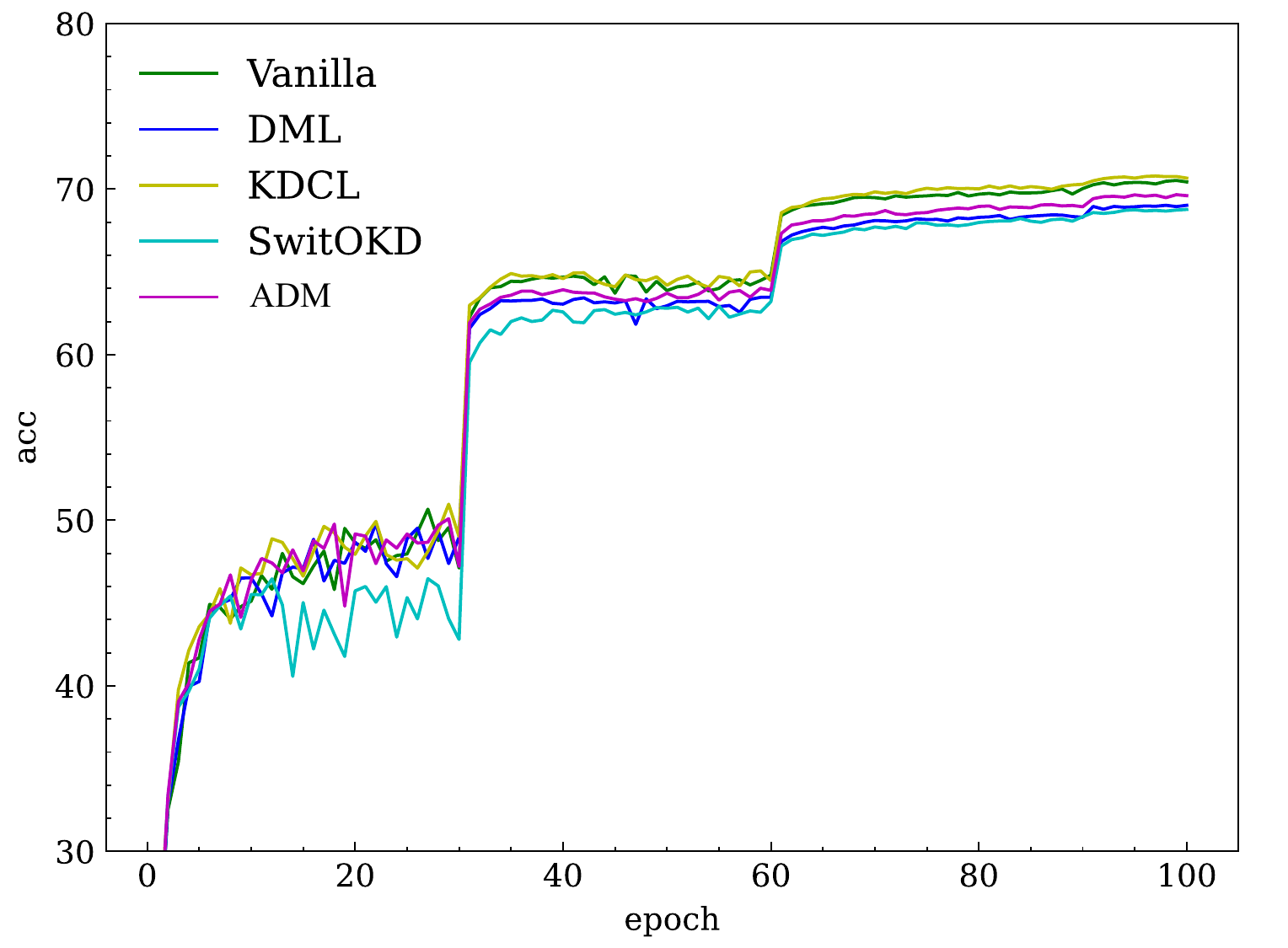}}
    \hspace{1mm}
    \subfigure[mean test acc]
    {\includegraphics[width=0.3\linewidth,height=0.21\linewidth]{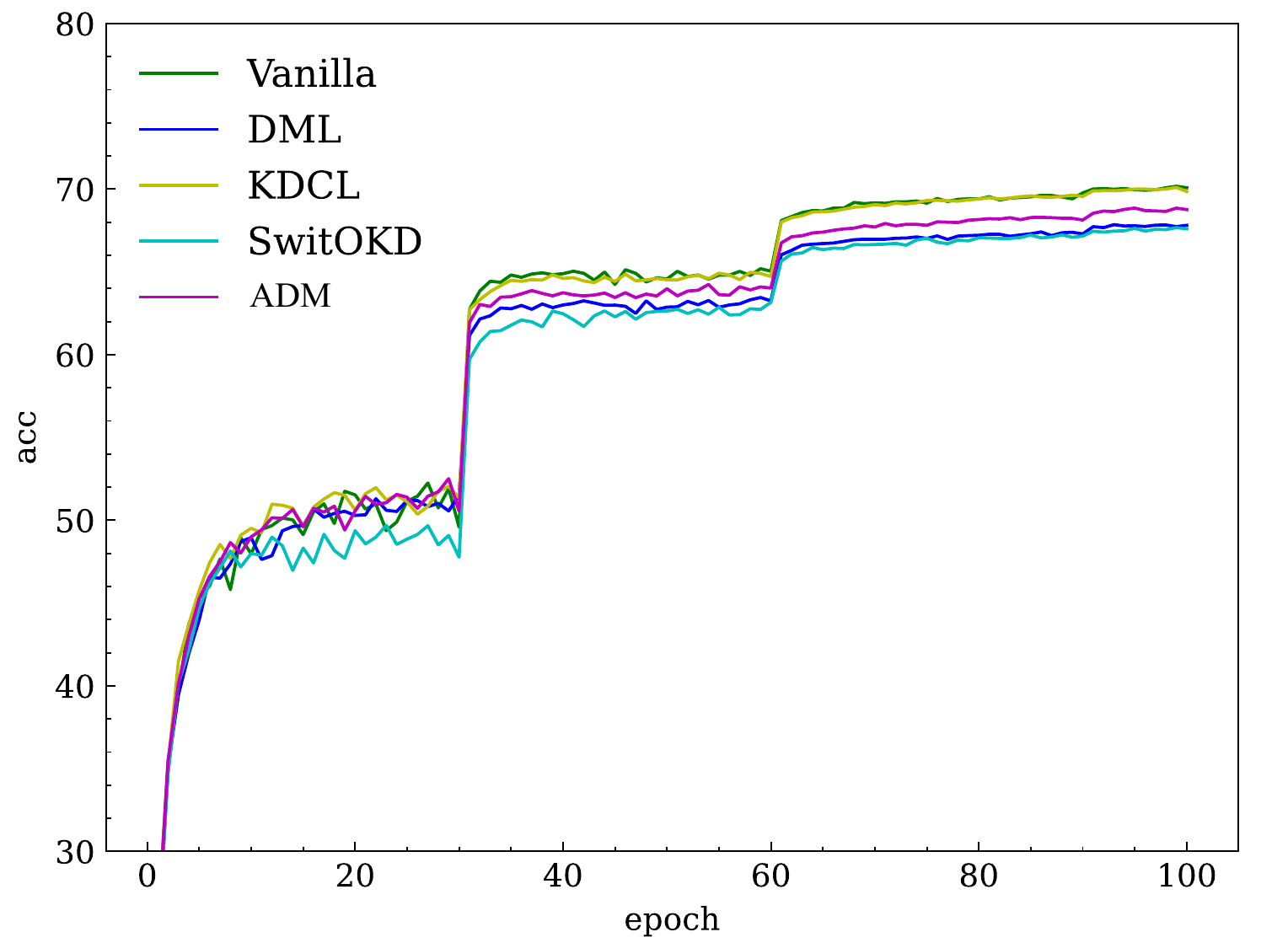}}
    \caption{\textbf{Testing Acc during training.} ResNet-18 $\rightarrow$ 0.5MobielNetV2 on ImageNet.
    }
    \label{fig:4}
\end{figure*}

As shown in Figure \ref{fig:3} and Figure \ref{fig:4}, We discover that ADM achieved superior performance on student's Accuracy (Acc), however, most OKD methods experienced a decline in performance for both the teacher model and mean prediction. A possible explanation is that current OKD approaches utilize the same distillation temperature (T), which imposes considerable constraints on the teacher model, consequently impacting the teacher model's performance.

\end{document}